%% file: main_neurips.tex
\documentclass[hidelinks]{article}
\input{preamble/main}

\usepgfplotslibrary{fillbetween}

\PassOptionsToPackage{numbers}{natbib}

      \usepackage[preprint]{neurips_2019}

\usepackage[utf8]{inputenc} 
\usepackage[T1]{fontenc}    
\usepackage{url}            
\usepackage{booktabs}       
\usepackage{amsfonts}       
\usepackage{nicefrac}       
\usepackage{microtype}      
\usepackage{caption}        
\usepackage{float}         
\usepackage{pbox}         
\usepackage{graphics}    

\title{Reconciling meta-learning and continual learning \\ with online mixtures of tasks}

\author{
    Ghassen Jerfel \footnotemark[1] \\
    \texttt{gj47@duke.edu} \\
    Duke University 
    \And
    Erin Grant \thanks{Equal contribution.} \\
    \texttt{eringrant@berkeley.edu} \\
    UC Berkeley
    \AND
    Thomas L. Griffiths \\
    \texttt{tomg@princeton.edu} \\
    Princeton University
    \And
    Katherine Heller \\
    \texttt{kheller@stat.duke.edu} \\
    Duke University
}

\begin{document}

\newcommand{\erin}[1]{({\color{orange}{Erin: #1})}}
\definecolor{turquoise}{rgb}{0.19, 0.84, 0.78}
\newcommand{\ghassen}[1]{({\color{turquoise}{Ghassen: #1})}}
\newcommand{\todo}[1]{({\color{red}{#1})}}
\newcommand{\annotation}[1]{%
  \marginpar{\small\itshape\color{blue}#1}}

\renewcommand{\erin}[1]{}
\renewcommand{\ghassen}[1]{}
\renewcommand{\todo}[1]{}
\renewcommand{\todocite}[1]{}

\newcommand{\pcol}[1]{{\color{black!20!blue}{#1}}}
\newcommand{\nonpcol}[1]{{\color{black!20!red}{#1}}}

\input{plot_defs}

\setlength{\interspacetitleruled}{0pt}%
\setlength{\algotitleheightrule}{0pt}%


\makeatletter
\renewcommand{\paragraph}{%
  \@startsection{paragraph}{4}%
  {\z@}{.5ex \@plus 1ex \@minus .2ex}{-1em}%
  {\normalfont\normalsize\bfseries}%
}
\makeatother

\maketitle

\begin{abstract}
    \input{abstract}
\end{abstract}

\input{sections/01_intro}

\vspace{-4pt}
\input{sections/02_background}

\vspace{-4pt}
\input{sections/03_param}

\input{figures/exp_miniimagenet}
\input{sections/04_exp_param}
\input{sections/05_nonparam}
\input{figures/alg_em_nonparametric}
\input{sections/06_exp_nonparam}
\input{figures/regression}
\vspace{-8pt}
\input{sections/07_literature_review}

\input{figures/classification}

\vspace{8pt}
\input{sections/08_conclusion}

\clearpage
\bibliographystyle{authordate1}
\bibliography{refs}

\clearpage
\appendix
\input{sections/09_supp}
\end{document}

%% file: preamble/main.tex
\usepackage[export]{adjustbox}
\usepackage{amsfonts}       
\usepackage{amsmath}        
\usepackage{amssymb}        
\usepackage{array}          
\usepackage{bm}             
\usepackage{booktabs}       
\usepackage{forest}         
\usepackage{graphicx}       
\usepackage{hyperref}       
\usepackage{ifthen}         
\usepackage{makecell}       
\usepackage{mathrsfs}       
\usepackage{microtype}      
\usepackage{multirow}       
\usepackage{pgfplots}       
\pgfplotsset{compat=1.14}
\usepackage{setspace}       %
\usepackage{subcaption}     
\usepackage{subdepth}       
\usepackage{tikz}           
\usepackage{verbatim}       
\usepackage{url}            
\usepackage{xfrac}          

\long\def\ignorethis#1{}

\newcommand{\eg}{\textit{e.g.,}~}
\newcommand{\ie}{\textit{i.e.,}~}
\newcommand{\cf}{\textit{c.f.,}~}
\newcommand{\vs}{\textit{vs.}~}

\definecolor{darkgreen}{RGB}{0,125,0}
\definecolor{orange}{RGB}{255,124,84}
\definecolor{turquoise}{rgb}{0.19, 0.84, 0.78}
\newcounter{NoteCounter}

\definecolor{shadecolor}{gray}{0.9}

\usepackage{graphicx}
\usepackage[font=small,labelfont=bf]{caption}
\usepackage[format=hang]{subcaption}

\usepackage[algo2e,algoruled,vlined,noend,ruled]{algorithm2e}  
\AtBeginEnvironment{algorithm2e}{\setstretch{1.05}}            
\SetAlFnt{\footnotesize}
\SetAlCapFnt{\footnotesize}
\SetAlCapNameFnt{\footnotesize}

\usepackage{listings}
\usepackage{fancyvrb}
\fvset{fontsize=\footnotesize}
\usepackage{etoolbox}

\usepackage[all]{hypcap}

\newcommand{\figref}[1]{\hyperref[fig:#1]{Figure~\ref{fig:#1}}}
\newcommand{\twofigref}[2]{Figures~\hyperref[fig:#1]{\ref{fig:#1}} and~\hyperref[fig:#2]{\ref{fig:#2}}}

\newcommand{\tabref}[1]{\hyperref[tab:#1]{Table~\ref{tab:#1}}}
\newcommand{\twotabref}[2]{Tables~\hyperref[tab:#1]{\ref{tab:#1}} and~\hyperref[tab:#2]{\ref{tab:#2}}}

\renewcommand{\eqref}[1]{\hyperref[eq:#1]{Equation~(\ref{eq:#1})}}
\newcommand{\shorteqref}[1]{\hyperref[eq:#1]{(\ref{eq:#1})}}
\newcommand{\twoeqref}[2]{Equations~\hyperref[eq:#1]{\ref{eq:#1}} and~\hyperref[eq:#2]{\ref{eq:#2}}}

\newcommand{\secref}[1]{\hyperref[sec:#1]{Section~\ref{sec:#1}}}
\newcommand{\twosecref}[2]{Sections~\hyperref[sec:#1]{\ref{sec:#1}} and~\hyperref[sec:#2]{\ref{sec:#2}}}

\newcommand{\appref}[1]{\hyperref[app:#1]{Appendix~\ref{app:#1}}}
\newcommand{\thmref}[1]{\hyperref[thm:#1]{Theorem~(\ref{thm:#1})}}
\newcommand{\algref}[1]{\hyperref[alg:#1]{Algorithm~\ref{alg:#1}}}
\renewcommand{\subref}[1]{\hyperref[alg:#1]{Subroutine~\ref{alg:#1}}}

\newcommand{\todocite}[1]{[\textcolor{blue}{#1}]}

\makeatletter
\let\UrlSpecialsOld\UrlSpecials
\def\UrlSpecials{\UrlSpecialsOld\do\/{\Url@slash}\do\_{\Url@underscore}}%
\def\Url@slash{\@ifnextchar/{\kern-.11em\mathchar47\kern-.2em}%
    {\kern-.0em\mathchar47\kern-.08em\penalty\UrlBigBreakPenalty}}
\def\Url@underscore{\nfss@text{\leavevmode \kern.06em\vbox{\hrule\@width.3em}}}
\makeatother

\usepackage[
  acronym,
  smallcaps,
  nowarn,
  section,
  nonumberlist
]{glossaries-extra}
\glsdisablehyper{}

\lstdefinestyle{mystyle}{
    commentstyle=\color{OliveGreen},
    numberstyle=\tiny\color{black!60},
    stringstyle=\color{BrickRed},
    basicstyle=\ttfamily\scriptsize,
    breakatwhitespace=false,
    breaklines=true,
    captionpos=b,
    keepspaces=true,
    numbers=none,
    numbersep=5pt,
    showspaces=false,
    showstringspaces=false,
    showtabs=false,
    tabsize=2
}
\tikzset{every picture/.style=remember picture}

\input{preamble/defs/acronyms}
\input{preamble/defs/math}

\input{preamble/defs/tikz}
\input{preamble/defs/project_specific}

%% file: preamble/defs/acronyms.tex
\setabbreviationstyle[long]{long}             
\setabbreviationstyle[long-noshort]{long-em}  

\newabbreviation[category=long-noshort]{ch}{}{cognitive hierarchy}
\newabbreviation[category=long-noshort]{da}{}{domain adaptation}
\newabbreviation[category=long-noshort]{eb}{}{empirical Bayes}
\newabbreviation[category=long-noshort]{es}{}{early stopping}
\newabbreviation[category=long-noshort]{fa}{}{fast adaptation}
\newabbreviation[category=long-noshort]{fisher}{}{Fisher information matrix}
\newabbreviation[category=long-noshort]{hbm}{}{hierarchical Bayesian model}
\newabbreviation[category=long-noshort]{ho}{}{hyperparameter optimization}
\newabbreviation[category=long-noshort]{iid}{}{independent and identically distributed}
\newabbreviation[category=long-noshort]{ib}{}{inductive bias}
\newabbreviation[category=long-noshort]{gb}{}{gradient-based}
\newabbreviation[category=long-noshort]{gen}{}{generative}
\newabbreviation[category=long-noshort]{joint}{}{joint training}
\newabbreviation[category=long-noshort]{jt}{}{joint training}
\newabbreviation[category=long-noshort]{kf}{}{Kronecker factor}
\newabbreviation[category=long-noshort]{laplace}{}{the Laplace approximation}
\newabbreviation[category=long-noshort]{ltl}{}{learning-to-learn}
\newabbreviation[category=long-noshort]{mse}{}{mean-squared error}
\newabbreviation[category=long-noshort]{mini}{}{\textit{mini}ImageNet}
\newabbreviation[category=long-noshort]{ml}{}{meta-learning}
\newabbreviation[category=long-noshort]{mnist}{}{MNIST}
\newabbreviation[category=long-noshort]{np}{}{non-parametric}
\newabbreviation[category=long-noshort]{omni}{}{Omniglot}
\newabbreviation[category=long-noshort]{p}{}{parametric}
\newabbreviation[category=long-noshort]{permmnist}{}{permuted MNIST}
\newabbreviation[category=long-noshort]{qfunc}{}{$Q$-function}
\newabbreviation[category=long-noshort]{rnn}{}{recurrent neural network}
\newabbreviation[category=long-noshort]{splitcifar}{}{split CIFAR-10/100}
\newabbreviation[category=long-noshort]{tl}{}{transfer learning}
\newabbreviation[category=long-noshort]{taskag}{}{\emph{task-agnostic}}
\newabbreviation[category=long-noshort]{taskaw}{}{\emph{task-aware}}
\newabbreviation[category=long-noshort]{tom}{}{theory of mind}
\newabbreviation[category=long-noshort]{vi}{}{variational inference}

\newabbreviation[category=long-short]{a3c}{A3C}{asynchronous advantage actor-critic}
\newabbreviation[category=long-short]{ban}{BAN}{born again network}
\newabbreviation[category=long-short]{bn}{BN}{batch normalization}
\newabbreviation[category=long-short]{bnp}{BNP}{Bayesian non-parametric}
\newabbreviation[category=long-short]{coma}{COMA}{counterfactual multi-agent}
\newabbreviation[category=long-short]{crp}{CRP}{Chinese restaurant process}
\newabbreviation[category=long-short]{dpmm}{DPMM}{Dirichlet process mixture model}
\newabbreviation[category=long-short]{dqn}{DQN}{deep Q-network}
\newabbreviation[category=long-short]{drl}{DRL}{deep reinforcement learning}
\newabbreviation[category=long-short]{em}{EM}{expectation maximization}
\newabbreviation[category=long-short]{ewc}{EWC}{elastic weight consolidation}
\newabbreviation[category=long-short]{gae}{GAE}{generalized advantage estimation}
\newabbreviation[category=long-short]{gbml}{GBML}{gradient-based meta-learning}
\newabbreviation[category=long-short]{hb}{HB}{hierarchical Bayes}
\newabbreviation[category=long-short]{icarl}{iCaRL}{incremental classifier and representation learning}
\newabbreviation[category=long-short]{icm}{ICM}{iterated conditional modes}
\newabbreviation[category=long-short]{ilsvrc}{ILSVRC}{ImageNet Large Scale Visual Recognition Challenge}
\newabbreviation[category=long-short]{ipomdp}{I-POMDP}{interactive Markov decision process}
\newabbreviation[category=long-short]{it}{IT}{iterated learning}
\newabbreviation[category=long-short]{kfac}{K-FAC}{Kronecker-factored approximate curvature}
\newabbreviation[category=long-short]{leo}{LEO}{latent embedding optimization}
\newabbreviation[category=long-short]{llama}{LLAMA}{lightweight Laplace approximation for meta-adaptation}
\newabbreviation[category=long-short]{lola}{LOLA}{learning with opponent-learning awareness}
\newabbreviation[category=long-short]{mle}{MLE}{maximum likelihood}
\newabbreviation[category=long-short]{map}{MAP}{\textit{maximum a posteriori}}
\newabbreviation[category=long-short]{maml}{MAML}{model-agnostic meta-learning}
\newabbreviation[category=long-short]{mppi}{MPPI}{model-predictive path integral}
\newabbreviation[category=long-short]{marl}{MARL}{multi-agent reinforcement learning}
\newabbreviation[category=long-short]{mdp}{MDP}{Markov decision process}
\newabbreviation[category=long-short]{mcmc}{MCMC}{Markov chain Monte Carlo}
\newabbreviation[category=long-short]{mn}{MN}{matching network}
\newabbreviation[category=long-short]{mpc}{MPC}{model predictive control}
\newabbreviation[category=long-short]{mt-net}{MT-net}{mask transformation network}
\newabbreviation[category=long-short]{nll}{NLL}{negative log likelihood}
\newabbreviation[category=long-short]{pomdp}{POMDP}{partially observable Markov decision process}
\newabbreviation[category=long-short]{platipus}{PLATIPUS}{Probabilistic LATent model for Incorporating Priors and Uncertainty in few-Shot learning)}
\newabbreviation[category=long-short]{rl}{RL}{reinforcement learning}
\newabbreviation[category=long-short]{rsa}{RSA}{rational speech acts}
\newabbreviation[category=long-short]{sgd}{SGD}{stochastic gradient descent}
\newabbreviation[category=long-short]{si}{SI}{synaptic intelligence}
\newabbreviation[category=long-short]{tadam}{TADAM}{task-dependent adaptive metric}
\newabbreviation[category=long-short]{td}{TD}{temporal difference}
\newabbreviation[category=long-short]{tf}{TF}{TensorFlow}
\newabbreviation[category=long-short]{vcl}{VCL}{variational continual learning}
\newabbreviation[category=long-short]{versa}{VERSA}{versatile amortized inference}

%% file: preamble/defs/math.tex
\newcommand{\smashinline}[2]{\mathchoice{#1}{#2}{#2}{#2}}

\newcommand{\bd}[1]{\partial{\kern0pt#1}}
\newcommand{\interior}[1]{{\kern0pt#1\kern1pt}^{\mathrm{o}}}

\newcommand{\softmaxwithtemperature}[1]{\tau\operatornamewithlimits{-softmax}_{#1}}
\newcommand{\assign}{\leftarrow}

\newcommand{\grad}[1]{\operatorname{\nabla}_{#1}}
\newcommand{\prob}[1]{p\brac{{\,}#1{\,}}}

\newcommand{\diff}{\mathop{}\!\mathrm{d}}

\newcommand{\tr}{^\mathrm{T}}

\newcommand{\sample}{\sim}
\newcommand{\given}{\;\vert\;}

\newcommand{\brac}[1]{\smashinline{\left(#1\right)}{(#1)}}
\newcommand{\cbrac}[1]{\smashinline{\left\{#1\right\}}{\{#1\}}}
\newcommand{\sbrac}[1]{\smashinline{\left[#1\right]}{[#1]}}

\newcommand{\boldvar}[1]{\bm{#1}}

\newcommand{\boldx}{\boldvar{x}}

\newcommand{\boldz}{\boldvar{z}}

\newcommand{\boldSigma}{\boldvar{\Sigma}}

\newcommand{\boldtheta}{\boldvar{\theta}}

\newcommand{\boldphi}{\boldvar{\phi}}

\newcommand{\boldmu}{\boldvar{\mu}}

\DeclareMathAlphabet{\mathsfit}{\encodingdefault}{\sfdefault}{m}{sl}
\SetMathAlphabet{\mathsfit}{bold}{\encodingdefault}{\sfdefault}{bx}{n}

\newcommand{\mean}{\boldmu}
\newcommand{\covariance}{\boldSigma}

\newcommand{\MVN}[3]{{\frac {\exp \left(-{\frac {1}{2}}({\mathbf {x} }-{\mean})\tr }{\covariance}^{-1}(\bx-{\mean})\right)}{\sqrt {|2\pi {\covariance|}}}}


%% file: preamble/defs/tikz.tex
\usepackage{pgfplots}
\usepackage{pgfplotstable}
\pgfplotsset{compat=1.3}
\usepackage{tikz}

\usetikzlibrary{matrix}
\usetikzlibrary{bayesnet}
\usetikzlibrary{calc}
\usetikzlibrary{decorations.pathmorphing}
\usetikzlibrary{positioning}
\usetikzlibrary{pgfplots.groupplots}
\usetikzlibrary{spy}

\pgfdeclarelayer{edgelayer}
\pgfdeclarelayer{nodelayer}
\pgfsetlayers{edgelayer,nodelayer,main}

\definecolor{grey}{rgb}{0.749,0.749,0.749}

\tikzset{>=latex}

\tikzstyle{tensor} = [circle,fill=white,draw=black]
\tikzstyle{op}     = [rectangle,fill=grey,draw=black]

%% file: preamble/defs/project_specific.tex
\newcommand{\clusterid}{\boldz}

\newcommand{\datapoint}{\boldx}

\newcommand{\temperature}{\tau}
\newcommand{\responsibility}{\gamma}
\newcommand{\concentration}{\zeta}
\newcommand{\weightedclusterindicator}{\rho}
\newcommand{\observationcount}{n}
\newcommand{\globalprior}{G_{0}}

\newcommand{\clusteridx}[1]{^{(#1)}}
\newcommand{\clusteridxvbl}{\ell}
\newcommand{\clustercount}{L}

\newcommand{\spawningthreshold}{\epsilon}

\newcommand{\metaparams}{\boldtheta}
\newcommand{\taskparams}{\boldphi}
\newcommand{\task}{\mathcal{T}}
\newcommand{\dataset}{\mathscr{D}}

\newcommand{\metalr}{\beta}
\newcommand{\fastlr}{\alpha}

\newcommand{\taskidx}[1]{_{#1}}
\newcommand{\taskidxvbl}{j}
\newcommand{\taskcount}{J}

\newcommand{\datapointidx}[1]{_{#1}}
\newcommand{\datapointidxvbl}{i}
\newcommand{\datapointcounttrn}{N}
\newcommand{\datapointcountval}{M}

\newcommand{\iteridxvbl}{k}
\newcommand{\itercount}{K}

%% file: plot_defs.tex
\newcommand*{\plotwidth}{0.8\textwidth}
\newcommand*{\plotheight}{0.12\textheight}
\newcommand{\plotaymin}{8}
\newcommand{\plotaymax}{18}
\newcommand{\plotbymin}{40}
\newcommand{\plotbymax}{70}
\newcommand{\plotcymin}{20}
\newcommand{\plotcymax}{50}

\pgfplotsset{
    continuallearning/.style={
        width=.95*\plotwidth, 
        height=\plotheight,
        legend style={font=\tiny,align=center},
        y label style={align=center,font=\bfseries\scriptsize},
        x label style={font=\bfseries\normalsize},
        tick label style={font=\footnotesize},
        scaled y ticks=false,
        scaled x ticks=false,
        set layers,
    },
}

\pgfkeys{%
    /tikz/on layer/.code={
        \pgfonlayer{#1}\begingroup
            \aftergroup\endpgfonlayer
            \aftergroup\endgroup
    }
}

\pgfplotsset{
    highlight/.code args={#1:#2}{
        \fill [every highlight] ({axis cs:#1,0}|-{rel axis cs:0,0}) rectangle ({axis cs:#2,0}|-{rel axis cs:0,1});
    },
        /tikz/every highlight/.style={
            on layer=\pgfkeysvalueof{/pgfplots/highlight layer},
            blue!20
        },
        /tikz/highlight style/.style={
            /tikz/every highlight/.append style=#1
        },
        highlight layer/.initial=axis background,
}

\pgfplotsset{
    unif-noreset/.style={
        mark=none, 
        color=blue,
    },
    unif-reset/.style={
        mark=none, 
        color=blue,
        dashed,
    },
    nonunif-noreset/.style={
        mark=none, 
        color=green,
    },
    nonunif-reset/.style={
        mark=none, 
        color=green,
        dashed,
    },
    nonparam/.style={
        mark=none, 
        color=red,
    },
    mt-net/.style={
        mark=none, 
        color=brown,
    },
    platipus/.style={
        mark=none, 
        color=green,
    },
    versa/.style={
        mark=none, 
        color=black!10!yellow,
    },
    switch/.style={
        mark=none, 
        black,
        highlight style={black!8},
        highlight layer=axis ticks,
    },
    cluster-1/.style={
        mark=none, 
        color=magenta,
    },
    cluster-2/.style={
        mark=none, 
        color=cyan,
    },
    cluster-3/.style={
        mark=none, 
        color=orange,
    },
}

%% file: abstract.tex
\emph{Learning-to-learn} or \emph{meta-learning} leverages data-driven inductive bias to increase the efficiency of learning on a novel task. This approach encounters difficulty when transfer is not advantageous, for instance, when tasks are considerably dissimilar or change over time. We use the connection between gradient-based meta-learning and hierarchical Bayes to propose a Dirichlet process mixture of hierarchical Bayesian models over the parameters of an arbitrary parametric model such as a neural network. In contrast to consolidating inductive biases into a single set of hyperparameters, our approach of task-dependent hyperparameter selection better handles latent distribution shift, as demonstrated on a set of evolving, image-based, few-shot learning benchmarks.

%% file: sections/01_intro.tex
\section{Introduction}
\label{sec:intro}

\Gls{ml} algorithms aim to increase the efficiency of learning 
by treating task-specific learning episodes as examples from which to
generalize~\citep{schmidhuber1987evolutionary}.
The central assumption of a meta-learning algorithm is that some tasks 
are inherently related and so inductive transfer
can improve sample efficiency and generalization
~\citep{caruana1993multitask,caruana1998multitask,baxter2000model}.
In learning a single set of domain-general hyperparameters that parameterize
a metric space~\citep{vinyals2016matching} or an optimizer~\citep{ravi2017optimization,finn2017model}, recent meta-learning
algorithms make the assumption that tasks are equally related, and therefore
non-adaptive, mutual transfer is appropriate.
This assumption has been cemented in recent few-shot learning benchmarks,
which comprise a set of tasks generated in a uniform
manner~\citep[\textit{e.g.,}][]{vinyals2016matching, finn2017model}.

However, the real world often presents scenarios in which an agent must decide
what degree of transfer is appropriate.
In some cases, a subset of tasks are more strongly related to each other, 
and so non-uniform transfer provides a strategic advantage.
On the other hand, transfer in the presence of dissimilar or outlier tasks
worsens generalization performance~\citep{rosenstein2005transfer,deleu2018effects}.
Moreover, when the underlying task distribution is non-stationary, inductive
transfer to previously observed tasks should exhibit graceful degradation to address
the catastrophic forgetting problem~\citep{kirkpatrick2017overcoming}.
In these settings, the consolidation of all inductive biases into a single set of
hyperparameters is not well-posed to deal with changing or diverse tasks.
In contrast, in order to account for this degree of task heterogeneity,
people detect and adapt to novel contexts
by attending to relationships between tasks~\citep{collins2013cognitive}.

In this work, we learn a mixture of hierarchical models
that allows a meta-learner to adaptively select over a set of
learned parameter initializations for \gls{gb} adaptation to a new task.
The method is equivalent to clustering task-specific parameters in the
hierarchical model induced by recasting \gls{gb} \gls{ml} as
hierarchical Bayes~\citep{grant2018recasting} and generalizes the \glsfirst{maml} algorithm
introduced in \cite{finn2017model}.
By treating the assignment of task-specific parameters to clusters
as latent variables, we can directly
detect similarities between tasks on the basis of the task-specific likelihood,
which may be parameterized by a complex model such as a neural network.
Our approach therefore alleviates the need for explicit geometric 
or probabilistic modeling assumptions about the weights
of a complex parametric model and provides a scalable method to regulate information transfer
between episodes.


We additionally consider the setting of a non-stationary or evolving task distribution,
which necessitates a meta-learning method that possesses adaptive complexity.
We extend our latent variable model to the \gls{np} setting and leverage
stochastic point estimation in an infinite mixture~\citep{rasmussen2000infinite}
over model parameters;
point estimation is scalable and requires no distributional assumptions, 
and is so the online gradient-based mixture approach is compatible with
any meta-learning algorithm or neural network architecture that admits gradient-based optimization.
We demonstrate the unexplored ability of this combination of \gls{np} parameter priors
with neural network models to automatically detect and adapt to task distribution shift in 
a naturalistic image dataset.
Our work tackles the non-trivial 
setting of \gls{taskag} continual learning--where the task
change is unobserved--thus addressing an unresolved challenge in 
\gls{taskaw} continual learning~\cite[\eg][]{kirkpatrick2017overcoming}.


%% file: sections/02_background.tex
\section{\large Gradient-based meta-learning as hierarchical Bayes}
\label{sec:background}

Since our approach is grounded in the probabilistic formulation of meta-learning as 
hierarchical Bayes~\citep{baxter1997bayesian},
our approach can be applied to any probabilistic meta-learner.
In this work, we focus on \gls{maml}~\citep{finn2017model},
a \gls{gb} \gls{ml} approach that estimates global parameters to be shared
among task-specific models as an initialization for
a few steps of gradient descent.
\Gls{maml} admits a natural interpretation as parameter estimation
in a hierarchical probabilistic model, where the learned initialization acts as data-driven
regularization for the estimation 
of task-specific parameters
$\smash{\hat{\taskparams}\taskidx{\taskidxvbl}}$.

In particular, \cite{grant2018recasting} cast \gls{maml} 
as posterior inference for task-specific parameters
$\taskparams\taskidx{\taskidxvbl}$ given some samples of task-specific data
$\datapoint\taskidx{\taskidxvbl\datapointidx{1:\datapointcounttrn}}$
and a prior over $\taskparams\taskidx{\taskidxvbl}$ 
that is induced by early stopping of an iterative optimization
procedure;
truncation at $K$ steps of gradient descent on the negative log-likelihood
\smash{
$-\log\prob{
  \datapoint\taskidx{\taskidxvbl\datapointidx{1:\datapointcounttrn}}
  \given
  \taskparams\taskidx{\taskidxvbl}
}$
}
starting from
\smash{$\taskparams\taskidx{\taskidxvbl_{(0)}} = \metaparams$}
can be then understood as mode estimation of the posterior
$\prob{
  \taskparams\taskidx{\taskidxvbl}
  \given
  \datapoint\taskidx{\taskidxvbl\datapointidx{1:\datapointcounttrn}},
  \metaparams
}~.$
The mode estimates
\smash{$\hat{\taskparams}\taskidx{\taskidxvbl} =
\taskparams\taskidx{\taskidxvbl_{(0)}}
    + \alpha 
\sum_{\iteridxvbl=1}^\itercount
\grad{\taskparams}
\log \prob{
    \datapoint\taskidx{\taskidxvbl\datapointidx{1:\datapointcounttrn}}
    \given 
    \taskparams\taskidx{\taskidxvbl_{(k-1)}}
}$}
are then combined to evaluate the marginal likelihood for each task as
\begin{align}
    \label{eq:marginal}
    \prob{
        \datapoint\taskidx{j\datapointidx{
            \datapointcounttrn\scalebox{0.5}[0.5]{+}1:\datapointcounttrn\scalebox{0.5}[0.5]{+}\datapointcountval}}
        \given
    \metaparams
    }
    & = \int 
    p\big(
       \datapoint\taskidx{j\datapointidx{
        \datapointcounttrn\scalebox{0.5}[0.5]{+}1:\datapointcounttrn\scalebox{0.5}[0.5]{+}\datapointcountval}}
    \given \taskparams\taskidx{\taskidxvbl}
    \big)
    p\big(
      \taskparams\taskidx{\taskidxvbl}
      \given \metaparams
    \big)
    \diff \taskparams\taskidx{\taskidxvbl}
    \approx 
    p \big(
       \datapoint\taskidx{j\datapointidx{
           \datapointcounttrn\scalebox{0.5}[0.5]{+}1:\datapointcounttrn\scalebox{0.5}[0.5]{+}\datapointcountval}}
      \given
      \hat{\taskparams\taskidx{\taskidxvbl}}
    \big)~,
\end{align}
where 
$\datapoint\taskidx{\taskidxvbl\datapointidx{\datapointcounttrn\scalebox{0.5}[0.5]{+}1:\datapointcounttrn\scalebox{0.5}[0.5]{+}\datapointcountval}}$ 
is another set of samples from the $j$th task.
A training dataset can then be summarized in an \gls{eb} point estimate of $\metaparams$ computed by
gradient-based optimization of the joint marginal likelihood in \shorteqref{marginal} in across tasks,
so that the likelihood of a datapoint sampled
from a new task can be computed using only $\metaparams$ and without storing the task-specific parameters.

%% file: sections/03_param.tex
\newcommand{\estep}{\texttt{E-STEP} }
\newcommand{\mstep}{\texttt{M-STEP} }

\section{Improving \gls{ml} by modeling latent task structure}
\label{sec:param}

If the task distribution is heterogeneous,
assuming a single parameter initialization
$\metaparams$ for gradient-based meta-learning is not suitable because
it is unlikely that the point estimate computed by a few steps
of gradient descent will sufficiently adapt the task-specific parameters
$\taskparams$ to a diversity of tasks.
Moreover, explicitly estimating relatedness between tasks has the potential to aid the efficacy of a
meta-learning algorithm by modulating both positive and negative
transfer~\citep{thrun1996discovering,zhang2010learning,rothman2010sparse,zhang2014regularization},
and by identifying outlier tasks that require a more significant degree of 
adaptation~\citep{xue2007multi,gupta2013factorial}.
Nonetheless, defining an appropriate notion of task relatedness
is a difficult problem in the high-dimensional parameter or activation space of models such as neural networks.

Using the probabilistic interpretation of \secref{background},
we deal with the variability in the tasks
by assuming that each set of task-specific parameters
$\taskparams\taskidx{\taskidxvbl}$
is drawn from a mixture of base distributions
each of which is parameterized by a hyperparameter
\smash{$\metaparams\clusteridx{\clusteridxvbl}$}.
Accordingly, we capture task relatedness by estimating the likelihood of
assigning each task to a mixture component
based simply on the task negative log likelihood after a few steps of
gradient-based adaptation.
The result is a scalable \gls{ml} algorithm that jointly
learns task-specific cluster assignments and model parameters, and is
capable of modulating the transfer of information across tasks by 
clustering together related task-specific parameter settings.

\input{figures/alg_em_maml}
\input{figures/alg_em_parametric}

Formally, let $\clusterid\taskidx{\taskidxvbl}$ be the categorical latent variable indicating the cluster
assignment of each task-specific parameter $\taskparams\taskidx{\taskidxvbl}$.
A direct maximization of the mixture model likelihood is a
gubernatorial optimization problem that can grow intractable.
This intractability is equally problematic for the posterior distribution over 
the cluster assignment variables $\clusterid\taskidx{\taskidxvbl}$ 
and the task-specific parameters $\taskparams\taskidx{\taskidxvbl}$, 
which are both treated as latent variables in the probabilistic formulation of meta-learning.
A scalable approximation involves representing the conditional distribution for each latent variable with a
\gls{map} estimate.
In our meta-learning setting of a mixture of \gls{hb} models, this suggests an augmented 
\gls{em} procedure~\citep{dempster1977maximum} alternating between an \estep that computes an expectation of the 
task-to-cluster assignments $\clusterid\taskidx{\taskidxvbl}$, which itself involves the computation of a conditional mode estimate for the task-specific parameters $\taskparams\taskidx{\taskidxvbl}$,
and an \mstep that updates the hyperparameters \smash{$\theta\clusteridx{1:\clustercount}$} (see Subroutine \ref{alg:em-mm}).

To ensure scalability, we use the minibatch variant of
stochastic optimization~\citep{robbins1951stochastic} 
in both the \estep and the \mstep;
such approaches to \gls{em} are motivated by a view of the
algorithm as optimizing a single free energy at both the \estep
and the \mstep~\citep{neal1998view}.
In particular, for each task $\taskidxvbl$ and cluster $\clusteridxvbl$,
we follow the gradients to minimize the negative log-likelihood on the
training data points 
$\smash{\datapoint\taskidx{j\datapointidx{1:\datapointcounttrn}}}$,
using the cluster parameters
$\smash{\metaparams\clusteridx{\clusteridxvbl}}$
as initialization.
This allows us to obtain a modal point estimate of the task-specific parameters
\smash{$\hat{\taskparams}\taskidx{\taskidxvbl}\clusteridx{\clusteridxvbl}$.}
The \estep
in Subroutine \ref{alg:em-mm}
leverages the connection
between \gls{gb} \gls{ml} and \gls{hb}~\citep{grant2018recasting}
and the differentiability of our clustering procedure
to employ the task-specific parameters to compute
the posterior probability of cluster assignment.
Accordingly, based on the likelihood of the same training data points under 
the model parameterized by
\smash{
  $\hat{\taskparams}\taskidx{\taskidxvbl}\clusteridx{\clusteridxvbl}$,
}
we compute the  cluster assignment probabilities as
\begin{align}
  \responsibility\taskidx{\taskidxvbl}\clusteridx{\clusteridxvbl}
  :=
  p\big(
    \clusterid\taskidx{\taskidxvbl} = \clusteridxvbl
    \given
    \datapoint\taskidx{j\datapointidx{1:\datapointcounttrn}},
    \metaparams\clusteridx{1:\clustercount}
    \big)
    \propto 
    \int
    p(
    \datapoint\taskidx{j\datapointidx{1:\datapointcounttrn}}
    \given
    \taskparams\taskidx{\taskidxvbl}
    )
    \ 
    p(
    \taskparams\taskidx{\taskidxvbl}
    \given
    \metaparams\clusteridx{\clusteridxvbl}
    )
    \diff \taskparams\taskidx{\taskidxvbl} 
    \approx
    p(
    \datapoint\taskidx{j\datapointidx{1:\datapointcounttrn}}
    \given
    \hat{\taskparams}\clusteridx{\clusteridxvbl}\taskidx{\taskidxvbl}
    )~.
\end{align}
The cluster means $\metaparams\clusteridx{\clusteridxvbl}$
are then updated by gradient descent on the validation loss
in the \mstep in Subroutine \ref{alg:em-mm};
this \mstep is analogous to the \gls{maml} algorithm in \cite{finn2017model} 
with the addition of mixing weights \smash{$\responsibility\taskidx{\taskidxvbl}\clusteridx{\clusteridxvbl}$}.

Note that, unlike other recent approaches to probabilistic
clustering~\citep[\eg][]{bauer2017discriminative}
we adhere to the episodic meta-learning setup for both
training and testing since only the task support set
$\datapoint\taskidx{\taskidxvbl\datapointidx{1:\datapointcounttrn}}$
is used to compute both the point estimate
\smash{
$\hat{\taskparams}\clusteridx{\clusteridxvbl}\taskidx{\taskidxvbl}$
}
and the cluster responsibilities
\smash{
$\responsibility\taskidx{\taskidxvbl}\clusteridx{\clusteridxvbl}$.
}
See \algref{em-maml} for the full algorithm, whose high-level structure is shared with the 
\gls{np} variant of our method detailed in \secref{nonparam}.

%% file: figures/alg_em_maml.tex
\newcommand{\indentvspace}{\vspace{1pt}}
\begin{figure}[t]
\centering
\setlength{\algomargin}{.5em}

\begingroup

\begin{algorithm2e}[H]

\SetInd{.5em}{.5em}

\SetKwFunction{algo}{Stochastic gradient-based EM for \pcol{finite} and \nonpcol{infinite} mixtures}
\SetKwFunction{estep}{E-STEP}
\SetKwFunction{mstep}{M-STEP}
\SetKwProg{myalg}{Algorithm 1}{}{}
\myalg{
  \algo{
      \\\,
    dataset $\dataset$,
    meta-learning rate $\metalr$,
    adaptation rate $\fastlr$,
    temperature $\temperature$,
    initial cluster count $\clustercount_0$,
    meta-batch size $\taskcount$,
    training batch size $\datapointcounttrn$,
    validation batch size $\datapointcountval$,
    adaptation iteration count $\itercount$,
    global prior $\globalprior$
  }
\indentvspace
\hrule
    \vspace{-5pt}
}{

\SetKwInOut{Input}{Input}
\SetKwInOut{Output}{Output}

Initialize cluster count
\pcol{$\clustercount \assign L_0$}
and meta-level parameters
\pcol{$\smash{\metaparams\clusteridx{1}, \dots,\metaparams\clusteridx{\clustercount} \sample \globalprior}$ }
\\

\While{not converged}{
  \indentvspace
    \indentvspace
  Draw tasks 
  \pcol{$\smash{\task\taskidx{1}, \dots, \task\taskidx{\taskcount}
    \sample p_\dataset\brac{\task}}$} \\
  \For{$\taskidxvbl$ in $1, \dots, \taskcount$}{
    \indentvspace
    \indentvspace
    Draw task-specific datapoints,
    \pcol{$\smash{\datapoint\taskidx{\taskidxvbl\datapointidx{1}}
    \dots
    \datapoint\taskidx{\taskidxvbl\datapointidx{\datapointcounttrn+\datapointcountval}}
    \sample
    p_{\task\taskidx{\taskidxvbl}}(\datapoint)}$} \\
      \indentvspace
    Draw a parameter initialization for a new cluster from the global prior,
    \nonpcol{$\metaparams\clusteridx{\clustercount+1} \sample \globalprior$} \\
    \For{$\clusteridxvbl$ in $\{1, \dots, \clustercount, \nonpcol{\clustercount+1}\}$}{
      \indentvspace
      \indentvspace
      Initialize
      \pcol{$\smash{\hat{\taskparams}\taskidx{\taskidxvbl}\clusteridx{\clusteridxvbl}
      \leftarrow
      \metaparams\clusteridx{\clusteridxvbl}}$}
      \indentvspace
      \\
      Compute task-specific mode  estimate,
      \pcol{$\smash{\hat{\taskparams}\taskidx{\taskidxvbl}\clusteridx{\clusteridxvbl}
      \leftarrow
      \hat{\taskparams}\taskidx{\taskidxvbl}\clusteridx{\clusteridxvbl}
      + \fastlr \sum_k \grad{\taskparams} 
      \log\prob{
        \datapoint\taskidx{\taskidxvbl\datapointidx{1:\datapointcounttrn}}
        \given
        \hat{\taskparams}\taskidx{\taskidxvbl}\clusteridx{\clusteridxvbl}
       }}$}\\
    }
    \indentvspace
    \indentvspace
    \indentvspace
    Compute assignment of tasks to clusters,
    \pcol{$\smash{\responsibility\taskidx{\taskidxvbl} \assign}$
    \estep
    $\smash{\brac{
    \datapoint\taskidx{
    \taskidxvbl\datapointidx{1:\datapointcounttrn}},
    \hat{\taskparams}\taskidx{\taskidxvbl}\clusteridx{1:\clustercount}
    }}$}
  }
  \indentvspace
  \indentvspace
  \indentvspace
  Update each component $\clusteridxvbl$ in $1, \dots, \clustercount$,
  \pcol{$\smash{\metaparams\clusteridx{\clusteridxvbl}
  \leftarrow\metaparams\clusteridx{\clusteridxvbl}} + $
  \mstep
  $\smash{(\cbrac{
  \datapoint\taskidx{\taskidxvbl\datapointidx{
  \datapointcounttrn\scalebox{0.5}[0.5]{+}1:\datapointcounttrn\scalebox{0.5}[0.5]{+}\datapointcountval}},
  \hat{\taskparams}\taskidx{\taskidxvbl}\clusteridx{\clusteridxvbl},
  \responsibility\taskidx{\taskidxvbl}
  }_{\taskidxvbl=1}^\taskcount)}$}
  \\
  \indentvspace
  Summarize \nonpcol{$\{\metaparams_1, \dots \}$ 
  to update global prior $\globalprior$}  \\
}

\indentvspace

\Return{ 
    $\cbrac{\metaparams\clusteridx{1},
    \dots }$
    }

}
{}
\end{algorithm2e}
\endgroup

\captionsetup{labelformat=empty}
\caption{}  
\label{alg:em-maml}

\vspace{-30pt}
\end{figure}

%% file: figures/alg_em_parametric.tex
\begin{figure}
\centering
\begin{minipage}{0.51\textwidth} 

\begingroup
    \begin{algorithm2e}[H]
    \SetKwFunction{proc}{E-STEP}
    \SetKwProg{myproc}{}{}{}
    \myproc{\proc{
    $\cbrac{\datapoint\taskidx{
    \taskidxvbl\datapointidx{\datapointidxvbl}}}_{
    \datapointidxvbl=1}^\datapointcounttrn,
    \cbrac{
    \hat{\taskparams}\taskidx{\taskidxvbl}\clusteridx{\clusteridxvbl}}_{
    \clusteridxvbl=1}^\clustercount$}
    }{
        \Return{    \pcol{$\softmaxwithtemperature{\clusteridxvbl}
    \brac{
      \sum_\datapointidxvbl
      \log\prob{
        \datapoint\taskidx{\taskidxvbl\datapointidx{\datapointidxvbl}}
        |
        \hat{\taskparams}\taskidx{\taskidxvbl}\clusteridx{\clusteridxvbl}
      }
  }$}
    }
    }{}
\end{algorithm2e} 
\endgroup
\end{minipage}
\begin{minipage}{0.48\textwidth} 
\begingroup
\begin{algorithm2e}[H]
    \SetKwFunction{proc}{M-STEP}
    \SetKwProg{myproc}{}{}{}
    \myproc{\proc{$ \cbrac{\datapoint\taskidx{
    \taskidxvbl\datapointidx{\datapointidxvbl}}}_{
    \datapointidxvbl=1}^\datapointcountval,
    \hat{\taskparams}\taskidx{\taskidxvbl}\clusteridx{\clusteridxvbl},
    \responsibility\taskidx{\taskidxvbl}
    $
    }}{
    \Return{
        \pcol{$\metalr  \grad{\metaparams}\sbrac{
    \sum_{\taskidxvbl,\datapointidxvbl}
    \responsibility\taskidx{\taskidxvbl}
    \log \prob{
        \datapoint\taskidx{\taskidxvbl\datapointidx{\datapointidxvbl}}
        |\,
        \hat{\taskparams}\taskidx{\taskidxvbl}\clusteridx{\clusteridxvbl}
      }
  }$}
    }
    }{}
\end{algorithm2e} 
\endgroup
\end{minipage}
\bgroup
\captionsetup{labelformat=empty}
\caption{
  Top: \textbf{Algorithm 1:}
  Stochastic \glsfirst{gb}
  \glsfirst{em} for probabilistic clustering of task-specific 
  parameters in a meta-learning setting.
  Bottom: \textbf{Subroutine 2:}
  The \texttt{E-STEP} and \texttt{M-STEP} for a finite 
  mixture of \gls{hbm}s implemented as gradient-based meta-learners.
\label{alg:em-mm}
}
\egroup
\vspace{-20pt}
\end{figure}

%% file: figures/exp_miniimagenet.tex
\newcommand{\notea}{\,\textsuperscript{a}\ }
\newcommand{\noteb}{\,\textsuperscript{b}\ }
\newcommand{\notec}{\,\textsuperscript{c}\ }
\begin{table}[!t]
\centering \footnotesize
\caption{
    Meta-test set accuracy on the \gls{mini} \textbf{5-way}, 
    \textbf{1-} and \textbf{5-shot} classification benchmarks from \cite{vinyals2016matching} among 
    methods using a comparable architecture (the 4-layer convolutional network
    from \cite{vinyals2016matching}).
    For methods on which we report results in later experiments, we additionally report the total number of parameters optimized by the meta-learning 
    algorithm.
    \scriptsize{
        \notea Results reported by \cite{ravi2017optimization}.
        \noteb We report test accuracy for models matching train and test ``shot'' and ``way''.
        \notec We report test accuracy for a comparable base (task-specific network) architecture.
}
\vspace{5pt}
}
\scalebox{.93}{
\begin{tabular}{p{180pt}p{52pt}p{14pt}p{1pt}p{16pt}p{14pt}p{1pt}p{18pt}}
\toprule
\textbf{Model}
& \textbf{Num. param.} 
& \multicolumn{3}{c}{\textbf{1-shot (\%)}} 
& \multicolumn{3}{c}{\textbf{5-shot} (\%)} \\
\midrule
     \textbf{matching network}   \tiny{\citep{vinyals2016matching}}\notea       &                & $43.56$ & $\pm$ & $0.84$ & $55.31$ & $\pm$ & $0.73$ \\
     \textbf{meta-learner LSTM}      \tiny{\citep{ravi2017optimization}}        &                & $43.44$ & $\pm$ & $0.77$ & $60.60$ & $\pm$ & $0.71$ \\
    \textbf{prototypical networks}  \tiny{\citep{snell2017prototypical}}\noteb &                & $46.61$ & $\pm$ & $0.78$ & $65.77$ & $\pm$ & $0.70$ \\
    \textbf{\acrshort{maml}}        \tiny{\citep{finn2017model}}               &                & $48.70$ & $\pm$ & $1.84$ & $63.11$ & $\pm$ & $0.92$ \\
    \textbf{\acrshort{mt-net}}       \tiny{\citep{lee2018gradient}}            & $38,907$ & $51.70$ & $\pm$ & $1.84$ &         &       &       \\
    \textbf{\acrshort{platipus}}    \tiny{\citep{finn2018probabilistic}}       & $65,546$ & $50.13$ & $\pm$ & $1.86$ &         &       &       \\
    \textbf{\acrshort{versa}}       \tiny{\citep{gordon2019meta}}\notec        & $807,938$& $48.53$ & $\pm$ & $1.84$ &       &       &       \\
\midrule                                                                     
    \textbf{Our method}: $2$ components                                        & $65,546$ & $49.60$ & $\pm$ & $1.50$  &  $64.60$  & $\pm$ & $0.92$  \\
    $3$ components                                                             & $98,319$ & $51.20$ & $\pm$ & $1.52$  &  $65.00$  & $\pm$ & $0.96$  \\
    $4$ components                                                             & $131,092$ & $50.49$ & $\pm$ & $ 1.46 $  & $64.78$  & $\pm$ & $1.43$ \\
    $5$ components                                                             & $163,865$ & $51.46$ & $\pm$ & $1.68$ &          &      &        \\
\bottomrule
\end{tabular}
}
\label{fig:miniimagenet}
    \vspace{-13pt}
\end{table}

%% file: sections/04_exp_param.tex
\section{Experiment: \gls{mini} few-shot classification}
\label{sec:exp-param}

Clustering task-specific parameters provides a way for a meta-learner to deal
with task heterogeneity as each cluster can be associated with
a subset of the tasks that would benefit most from mutual transfer.
While we do not expect existing tasks to present a significant degree of
heterogeneity given the uniform sampling assumptions behind their design,
we nevertheless conduct an experiment to validate that our method gives an improvement on a standard
benchmark for few-shot learning.

We apply Algorithm \ref{alg:em-dpmm} with Subroutine \ref{alg:em-mm} and $L\in\{2, 3, 4, 5\}$ components to the 1-shot 
and 5-shot, 5-way, \gls{mini} few-shot classification benchmarks~\citep{vinyals2016matching};
Appendix \ref{sec:supp-classification} contains additional experimental details.
We demonstrate in Table 1 that a mixture of meta-learners improves the performance
of gradient-based meta-learning on this task for any number of components.
However, the performance of the parametric mixture does not improve monotonically with the 
number of components $L$.
This leads us to the development of \gls{np} clustering for continual meta-learning,
where enforcing specialization to subgroups of tasks and
increasing  model complexity is in fact necessary to preserve performance on prior tasks
due to significant heterogeneity.

%% file: sections/05_nonparam.tex
\section{Scalable online mixtures for task-agnostic continual learning}
\label{sec:nonparam}

The mixture of meta-learners developed in \secref{param} addresses a drawback of meta-learning
approaches such as \gls{maml} that consolidate task-general information into a single set of hyperparameters.
However, the method adds another dimension to model selection in the form of identifying the correct number of mixture components.
While this may be resolved by cross-validation if the dataset is static 
and therefore the number of components can remain fixed, 
adhering to a fixed number of components throughout training 
is not appropriate in the non-stationary regime,
where the underlying task distribution changes as different 
types of tasks are presented sequentially in a continual learning setting.
In this regime, it is important to incrementally introduce more components 
that can each specialize to the distribution of tasks observed at the time of spawning.

To address this, we derive a scalable stochastic estimation procedure to 
compute the expectation of task-to-cluster assignments (\estep)
for a growing number of task clusters in a \emph{\gls{np}} mixture 
model~\citep{rasmussen2000infinite} called the \glsfirst{dpmm}.
The formulation of the \gls{dpmm} that is most appropriate for 
incremental learning is the sequential draws formulation that corresponds to an instantiation 
of the \gls{crp}~\citep{rasmussen2000infinite}.
A \gls{crp} prior over $\clusterid\taskidx{\taskidxvbl}$ allows some probability to be assigned to a 
new mixture component while the task identities are inferred in a sequential manner,
and has therefore been key to recent online and stochastic learning 
of the \gls{dpmm}~\citep{lin2013online}.
A draw from a \gls{crp} proceeds as follows:
For a sequence of tasks, the first task is assigned to
the first cluster and the $\taskidxvbl$th subsequent task is then assigned to the 
$\clusteridxvbl$th cluster with probability
\begin{align}
  \prob{
    \clusterid\taskidx{\taskidxvbl} = \clusteridxvbl
    \given
    \clusterid\taskidx{1:\taskidxvbl-1},
  \concentration
  } =
  \left\{
    \begin{array}{ll}
      \observationcount\clusteridx{\clusteridxvbl}/\observationcount + \concentration
        & \textrm{for} \, \clusteridxvbl \leq L\\[.5em]
      \concentration/\observationcount + \concentration
      & \textrm{for} \, \clusteridxvbl = \clustercount + 1
      ~,
    \end{array}
  \right.
\end{align}
where $\clustercount$ is the number of non-empty clusters,
$\observationcount\clusteridx{\clusteridxvbl}$ is the number of tasks
already occupying a cluster $\clusteridxvbl$,
and $\concentration$ is a fixed positive concentration parameter.
The prior probability associated with a new mixture component
is therefore 
$\prob{
  \clusterid\taskidx{\taskidxvbl} = \clustercount + 1
\given
\clusterid\taskidx{1:\taskidxvbl-1},
\concentration
}$.

In a similar spirit to \secref{param}, we develop a stochastic \gls{em} procedure for the estimation
of the latent task-specific parameters \smash{$\taskparams\taskidx{1:\taskcount}$} and the meta-level 
parameters \smash{$\metaparams\clusteridx{1:\clustercount}$} in the \gls{dpmm}, which
allows the number of observed task clusters to grow in an online manner with the diversity of the task distribution.
While computation of the mode estimate of the task-specific parameters $\taskparams\taskidx{\taskidxvbl}$
is mostly unchanged from the finite variant, the estimation of the cluster assignment variables $\clusterid$ 
in the \estep requires revisiting 
the Gibbs conditional distributions due to the potential addition of a new cluster at each step.
For a \gls{dpmm}, the conditional distributions for $\clusterid\taskidx{\taskidxvbl}$ are
\begin{align}
  \prob{
    \clusterid\taskidx{\taskidxvbl} = \clusteridxvbl
    |\ 
    \datapoint\taskidx{j\datapointidx{1:\datapointcountval}},
    \clusterid\taskidx{1:\taskidxvbl-1}
  } 
  \propto
  \left\{
    \begin{array}{ll}
        \observationcount\clusteridx{\clusteridxvbl}{\int} 
        p( \datapoint\taskidx{j\datapointidx{1:\datapointcountval}}
            | \taskparams\taskidx{\taskidxvbl}\clusteridx{\clusteridxvbl}) 
        p( \taskparams\taskidx{\taskidxvbl}\clusteridx{\clusteridxvbl} | \metaparams)
        \,  \mathrm{d} \taskparams\taskidx{\taskidxvbl}
        \,  \mathrm{d} G_\clusteridxvbl(\metaparams)
        & \textrm{for} \ \clusteridxvbl \leq \clustercount\\[0.5em]
        \concentration \int 
        p( \datapoint\taskidx{j\datapointidx{1:\datapointcountval}}
            | \taskparams\taskidx{\taskidxvbl}\clusteridx{0}) 
        p( \taskparams\taskidx{\taskidxvbl}\clusteridx{0} | \metaparams)
        \,  \mathrm{d} \taskparams\taskidx{\taskidxvbl}
        \,  \mathrm{d} G_0(\metaparams)
        & \textrm{for} \ \clusteridxvbl = \clustercount + 1
    \end{array}
  \right.
  \label{eq:crp-prior}
\end{align}
with $\globalprior$ as the base measure or global prior over the components of the \gls{crp},
$G_\clusteridxvbl$ is the prior over each
cluster's parameters, initialized with a draw from a Gaussian centered at $\globalprior$ with a fixed variance
and updated over time using summary statistics from the set of active components 
\smash{$\{\metaparams\clusteridx{0}, \dots, \metaparams\clusteridx{\clustercount}\}$}.

Taking the logarithm of the posterior over task-to-cluster assignments $\clusterid\taskidx{\taskidxvbl}$ in (\ref{eq:crp-prior}) 
and using a mode estimate \smash{$\hat{\taskparams}\taskidx{\taskidxvbl}\clusteridx{\clusteridxvbl}$}
for task-specific parameters \smash{$\taskparams\taskidx{\taskidxvbl}$} as drawn from the $\clusteridxvbl$th cluster
gives the \estep in Subroutine \ref{alg:em-dpmm}.
We may also omit the prior term 
\smash{$\log \prob{\hat{\taskparams}\taskidx{\taskidxvbl}\clusteridx{\clusteridxvbl} \given \metaparams\clusteridx{\clusteridxvbl}}$}
as it arises as an implicit prior resulting from truncated gradient descent, 
as explained in \secref{param} of~\cite{grant2018recasting}.


%% file: figures/alg_em_nonparametric.tex
\begin{figure}[!t]
\centering
    \begin{algorithm2e}[H]
    \SetKwFunction{proc}{E-STEP}
    \SetKwProg{myproc}{}{}{}
    \myproc{\proc{
        $\datapoint\datapointidx{\taskidxvbl_{1:\datapointcounttrn}},
    \hat{\taskparams}\taskidx{\taskidxvbl}\clusteridx{1:\clustercount},$
    \text{concentration} $\concentration$,
    \text{threshold} $\spawningthreshold$
    }}{
      \acrshort{dpmm} log-likelihood for all $\clusteridxvbl$ in $1, \dots, \clustercount$,
      \nonpcol{
      $\weightedclusterindicator\taskidx{\taskidxvbl}\clusteridx{\clusteridxvbl}
      \assign
      \sum_{\datapointidxvbl}
      \log\prob{
        \datapoint\taskidx{\taskidxvbl\datapointidx{\datapointidxvbl}}
        \given
        \hat{\taskparams}\taskidx{\taskidxvbl}\clusteridx{\clusteridxvbl}
      }
        + \log \observationcount\clusteridx{\clusteridxvbl}$}
      \\
      \acrshort{dpmm} log-likelihood
      for new component,
      \nonpcol{
      $\weightedclusterindicator\taskidx{\taskidxvbl}\clusteridx{\clustercount+1}
      \assign
      \sum_{\datapointidxvbl}
      \log \prob{
        \datapoint\taskidx{\taskidxvbl\datapointidx{\datapointidxvbl}}
        \given
        \hat{\taskparams}\taskidx{\taskidxvbl}\clusteridx{\clustercount+1}
      }
        + \log \concentration$}
      \\
      DPMM assignments,
      \nonpcol{
      $\responsibility\taskidx{\taskidxvbl}
      \leftarrow
      \softmaxwithtemperature{}
      \brac{
        \weightedclusterindicator\taskidx{\taskidxvbl}\clusteridx{1},
        \dots,
        \weightedclusterindicator\taskidx{\taskidxvbl}\clusteridx{\clustercount +1}
    }$} \\
      \uIf{$\responsibility\taskidx{\taskidxvbl}\clusteridx{\clustercount+1}
        > \spawningthreshold$}{
        Expand the model by incrementing \nonpcol{$\clustercount \leftarrow \clustercount
        + 1$} \\
      }
      \uElse{
        Renormalize \nonpcol{$\responsibility\taskidx{\taskidxvbl}
        \leftarrow
        \softmaxwithtemperature{}
        \brac{
          \weightedclusterindicator\taskidx{\taskidxvbl}\clusteridx{1},
          \dots,
          \weightedclusterindicator\taskidx{\taskidxvbl}\clusteridx{\clustercount}
      }$}
      }
      \Return{
      $\responsibility\taskidx{\taskidxvbl}$
    }
    }{}
\end{algorithm2e}

\begingroup
\begin{algorithm2e}[H]
    \SetKwFunction{proc}{M-STEP}
    \SetKwProg{myproc}{}{}{}
    \myproc{\proc{
    $
    \cbrac{\datapoint\taskidx{
    \taskidxvbl\datapointidx{\datapointidxvbl}}}_{
    \datapointidxvbl=1}^\datapointcountval,
    \hat{\taskparams}\taskidx{\taskidxvbl}\clusteridx{\clusteridxvbl},
    \responsibility\taskidx{\taskidxvbl},$
    \text{concentration} $\concentration$
    }}{
    \Return{
        \nonpcol{$\metalr \, \grad{\metaparams} \sbrac{
    \sum_{\taskidxvbl,\datapointidxvbl}
    \responsibility\taskidx{\taskidxvbl}
    \log \prob{
      \datapoint\taskidx{\taskidxvbl\datapointidx{\datapointidxvbl}}
      \given
      \hat{\taskparams}\taskidx{\taskidxvbl}\clusteridx{\clusteridxvbl}
    }
    + \log \observationcount\clusteridx{\clusteridxvbl}}
    $
    }
    }}
    {}
\end{algorithm2e} 
\endgroup
\vspace{-5pt}

\bgroup
\renewcommand{\figurename}{Subroutine}
\caption{
    The \texttt{E-STEP} and \texttt{M-STEP} for an infinite 
  mixture of \gls{hbm}s.
  \vspace{-15pt}
}
\label{alg:em-dpmm}
\egroup
\end{figure}

%% file: sections/06_exp_nonparam.tex
\section{Experiments: \Gls{taskag} continual few-shot regression \& classification}
\label{sec:exp-nonparam}

By treating the assignment of tasks to clusters as latent variables, 
the algorithm of \secref{nonparam} can adapt 
to a changing distribution of tasks, without any external information to signal distribution shift
(\ie in a \gls{taskag} manner).
Here, we present our main experimental results on both
a novel synthetic regression benchmark
as well as a novel evolving variant of \gls{mini}, and confirm the algorithm's
ability to adapt to distribution shift by spawning a 
newly specialized cluster.


\paragraph{High-capacity baselines.}
%
As an ablation, we compare to the \textbf{non-uniform} parameteric \textbf{mixture} proposed in \secref{param} with
the number of components fixed at the total number of task distributions in the dataset ($3$).
We also consider a \textbf{uniform} parametric \textbf{mixture} in which each component receives equal assignments;
this can also be seen as the non-uniform mixture in the infinite temperature ($\temperature$) limit.
%
Note that our meta-learner has a lower capacity than these two baselines for most of the
training procedure, as it may decide to expand its capacity past one component only when the 
task distribution changes.
Finally, for the large-scale experiment in 
\secref{continual-classification},
we compare with three recent meta-learning algorithms that report 
improved performance on the standard \gls{mini} benchmark of \secref{param},
but are not explicitly posed to address the continual learning setting of evolving tasks:
\textbf{\acrshort{mt-net}}~\citep{lee2018gradient}, 
\textbf{\acrshort{platipus}}~\citep{finn2018probabilistic}, 
and \textbf{\acrshort{versa}}~\citep{gordon2019meta}.

\input{figures/datasets}

\subsection{Continual few-shot regression}
\label{sec:continual-regression}

We first consider an explanatory experiment in which three regression tasks 
are presented sequentially with no overlap.
For input $\datapoint$ sampled uniformly from $[-5, 5]$, each regression task is
generated, in a similar spirit to the sinusoidal regression setup in \cite{finn2017model},
from one of a set of simple but distinct one-dimensional functions
(polynomial~\figref{polynomial},
sinusoid wave~\figref{sinusoid}, and
sawtooth wave~\figref{sawtooth}).
For the experiment in \figref{regression-loss} and \figref{regression-weight}, 
we presented the polynomial tasks for $4000$ iterations, 
followed by sinusoid tasks for $3000$ iterations,
and finally sawtooth tasks.
Additional details on the experimental setup can be found in Appendix \ref{sec:supp-continual-regression}.

\paragraph{Results: Distribution shift detection.}
The cluster responsibilities in \figref{regression-weight} on the meta-test dataset of tasks, 
from each of the three regression types in \figref{regressions},
indicates that the \gls{np} algorithm recognizes a change in the task 
distribution and spawns a new cluster at iterations $4000$ and a bit after $7000$.
Each newly created cluster is specialized to the task distribution observed at the 
time of spawning and remains as such throughout training, since
the majority of assignments for each type of regression
remains under a given cluster from the time of its introduction.

\paragraph{Results: Improved generalization and slower degradation of performance.}
We investigate the progression of the meta-test mean-squared error (MSE) for 
the three regression task distributions in \figref{regression-loss}.
We first note the clear advantage of non-uniform assignment both in improved generalization, 
when testing on the active task distribution,
and in slower degradation, 
when testing on previous distributions. 
This is due to the ability of these methods to modulate the transfer of 
information in order to limit negative transfer.
In contrast, the uniform method cannot selectively adapt specific clusters to be responsible for any given task,
and thus inevitably suffers from catastrophic forgetting.

The adaptive capacity of our \gls{np} method allows it to spawn clusters that specialize to newly observed tasks. 
Accordingly, even if the overall capacity is lower than that of the comparable non-uniform parametric method, 
our method achieves similar or better generalization, at any given training iteration.
More importantly, specialization allows our method to better modulate information transfer 
as the clusters are better differentiated. Consequently, each cluster does
not account for many assignments from more than a single task distribution, throughout training.
Therefore, we observed a significantly slower rate of degradation of the MSE on previous task distributions 
as new tasks are introduced. This is especially evident from the performance on the first task  in \figref{regression-loss}.

\subsection{Continual few-shot classification}
\label{sec:continual-classification}

Next, we consider an evolving variant of the \gls{mini} few-shot
classification task.
In this variant, one of a set of artistic filters are applied to the images during the
meta-training procedure to simulate a changing distribution of few-shot classification tasks.
For the experiment in Figure 8 and Figure 9 we first train using images with a 
``blur'' filter~(\figref{blur}) for $7500$ iterations,
then with a ``night'' filter~(\figref{night}) for another $7500$ iterations, 
 and finally with a ``pencil'' filter~(\figref{pencil}).
%
%
Additional details on the experimental setup can be found in Appendix \ref{sec:supp-continual-classification}.

\paragraph{Results: Meta-test accuracy.}
In Figure 9, we report the evolution of the meta-test accuracy for two variants of our
\gls{np} meta-learner in comparison to the parametric baselines introduced in \secref{exp-nonparam}, \textit{high-capacity baselines}.
The \gls{taskag} variant is the core algorithm described in previous sections, as used for the regression tasks.
The \gls{taskaw} variant augments the core algorithm with a cool-down period that prevents overspawning for the duration of a training phase. This requires some knowledge of the duration which is external to the meta-learner, thus the \gls{taskaw} nomenclature
(see Appendix \ref{sec:practical} for further details).

It is clear from Figure 8 that neither of our algorithms suffer from catastrophic 
forgetting to the same degree as the parametric baselines.
In fact, at the end of training, both of our methods outperform all the parametric baselines on the first and second task.

\paragraph{Results: Specialization.}
Given the higher capacity of the parametric baselines,
and the inherent degree of similarity between the filtered \gls{mini} task distributions
(unlike the regression tasks in the previous section),
the parametric baselines perform better on each task distribution while 
during its active phase.
However, they quickly suffer from degradation once the task distribution shifts.
Our approach does not suffer from this phenomenon and can handle non-stationarity owing
to the credit assignment of a single task distribution to a specialized cluster.
This specialization is illustrated in Figure 9, where we track the evolution of the average cluster responsibilities 
on the meta-test dataset from each of the three \gls{mini} few-shot classification tasks.
Each cluster is specialized so as to acquire the majority of a single task distribution's test set assignments,
despite the degree of similarity between tasks originating from the same source (\gls{mini}).
We observed this difficulty with the non-monotone improvement of parametric clustering, as a function of components, in Section 4.

%% file: figures/datasets.tex
\begin{figure}[!t]
\begin{minipage}[!t]{.46\columnwidth}
\begin{figure}[H]

\begin{subfigure}[b]{.3\columnwidth}
\begin{tikzpicture}
\begin{axis}[
    height=.8\columnwidth,
    width=\columnwidth,
    grid=major,
    hide axis, scale only axis,
]

\addplot[scatter,only marks,mark options={mark size=1, line width=1pt}] file [skip first] {data/datasets/polynomial.csv};
\end{axis}
\end{tikzpicture}
\caption{polynomial}
\label{fig:polynomial}
\end{subfigure}
\hskip1em
\begin{subfigure}[b]{.25\columnwidth}
\begin{tikzpicture}
\begin{axis}[
    height=.8\columnwidth,
    width=\columnwidth,
    grid=major,
    hide axis, scale only axis,
]

\addplot[scatter,only marks,mark options={mark size=1, line width=1pt}] file [skip first] {data/datasets/sinusoid.csv};
\end{axis}
\end{tikzpicture}
\caption{sinusoid}
\label{fig:sinusoid}
\end{subfigure}
\hskip1em
\begin{subfigure}[b]{.25\columnwidth}
\begin{tikzpicture}
\begin{axis}[
    height=.8\columnwidth,
    width=\columnwidth,
    grid=major,
    hide axis, scale only axis,
]

\addplot[scatter,only marks,mark options={mark size=1, line width=1pt}] file [skip first] {data/datasets/sawtooth.csv};
\end{axis}
\end{tikzpicture}
\caption{sawtooth}
\label{fig:sawtooth}
\end{subfigure}

\caption{
    The diverse set of periodic functions used for few-shot regression
    in \secref{continual-regression}.
}
\label{fig:regressions}
\end{figure}
\end{minipage}
\hfill
\begin{minipage}[!t]{.5\columnwidth}
\begin{figure}[H]
\begin{subfigure}[b]{.2\columnwidth}
\includegraphics[width=\columnwidth]{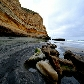}
\caption{plain}
\label{fig:unfiltered}
\end{subfigure}
\hskip1em
\begin{subfigure}[b]{.2\columnwidth}
\includegraphics[width=\columnwidth]{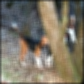}
\caption{blur}
\label{fig:blur}
\end{subfigure}
\hskip1em
\begin{subfigure}[b]{.18\columnwidth}
    \includegraphics[width=\columnwidth]{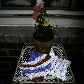}
    \caption{night}
    \label{fig:night}
\end{subfigure}
\hskip1em
\begin{subfigure}[b]{.2\columnwidth}
\includegraphics[width=\columnwidth]{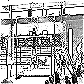}
\caption{pencil}
\label{fig:pencil}
\end{subfigure}
\caption{
    Artistic filters \textbf{(b-d)} applied to \gls{mini} \textbf{(a)} to
    ensure non-homogeneity of tasks in \secref{continual-classification}.
}
\label{fig:filters}
\end{figure}
\end{minipage}
\vspace{-10pt}
\end{figure}

%% file: figures/regression.tex
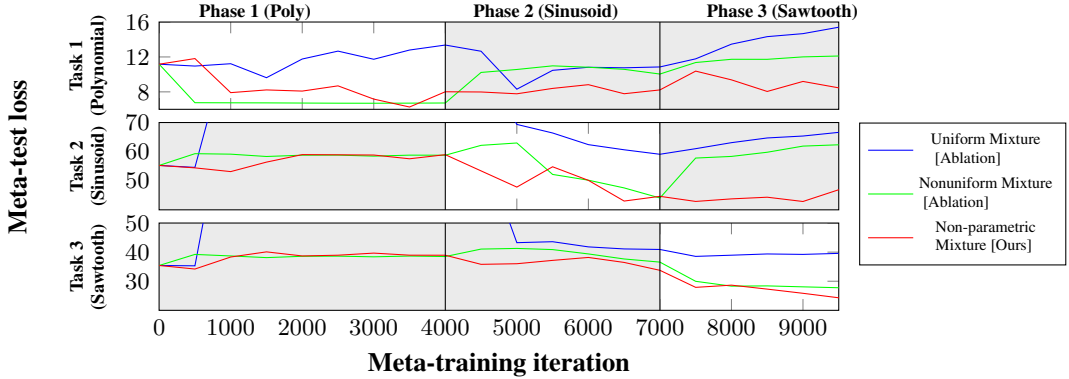
\begin{figure*}[t]
  \centering
  \input{figures/psSW-loss}
  \vspace{-6pt}
  \caption{
      Results on the evolving dataset of few-shot regression tasks (lower is better).
    Each panel (row) presents, for a specific task type (polynomial, sinusoid or sawtooth), 
    the average meta-test set accuracy of each method over cumulative number of few-shot episodes.
    We additionally report the degree of loss in backward transfer (i.e., catastrophic forgetting) to the tasks in each meta-test set in the legend;
    all methods but the \gls{np} method experience a large degree of catastrophic forgetting during an inactive phase.
  }
  \label{fig:regression-loss}
\vspace{-10pt}
\end{figure*}
\begin{figure*}[t]
    \centering
    \input{figures/psSW-weight}
    \vspace{-6pt}
    \caption{
        Each panel (row) presents task-specific per-cluster meta-test responsibilities $\gamma^{(\ell)}$ 
        over time. A higher responsibility entails a higher degree of specialization of 
        a particular cluster (color) to a particular task (row).
    }
    \label{fig:regression-weight}
\vspace{-12pt}
\end{figure*}

%% file: figures/psSW-loss.tex
\begin{tikzpicture}[baseline]


\begin{groupplot}[
      continuallearning,
      group style={group size=1 by 3, vertical sep=.5em},
      xmin=-1, xmax=9500,
      restrict y to domain*=0:100,  
      xticklabels={,,},
    ]

    \nextgroupplot[
        ylabel={Task 1 \\ (Polynomial)},
        ymin=\plotaymin, ymax=\plotaymax,
        ytick={10,14,18},
        yticklabels={$8$,$12$,$16$},
        legend pos=outer north east,
        xlabel={\hspace{20pt} \scriptsize{Phase 1 (Poly)} \hspace{58pt} \scriptsize{Phase 2 (Sinusoid)} \hspace{30pt} \scriptsize{Phase 3 (Sawtooth)}},
        xlabel shift={-50pt},
    ]

    \addplot+[unif-noreset]       table [x=itr, y=t1, col sep=comma, color=blue] {data/new-toy/unif-reset-slow-AvgLossB600.csv};
    \addplot+[nonunif-noreset]    table [x=itr, y=t1, col sep=comma, color=green] {data/new-toy/nonunif-reset-slow-WeightedLossB600.csv};
    \addplot+[nonparam]         table [x=itr, y=t1, col sep=comma, color=brown] {data/new-toy/nonp-WeightedLossB-Adj600.csv};

    \addplot+[switch, highlight=4000:9500]  coordinates {(4000, \plotaymin) (4000, \plotaymax)};
    \addplot+[switch]                        coordinates {(7000, \plotaymin) (7000, \plotaymax)};

    \nextgroupplot[
        ylabel={Task 2 \\ (Sinusoid)},
        legend pos=outer north east,
        ymin=\plotbymin, ymax=\plotbymax,
        legend pos=outer north east,
        legend style={align=left},
        ytick={50,60,70},
    ]
    \addplot+[unif-noreset]       table [x=itr, y=t2, col sep=comma, color=blue] {data/new-toy/unif-reset-slow-AvgLossB600.csv};
    \addlegendentry{Uniform Mixture \\ ~[Ablation]}
    \addplot+[nonunif-noreset]    table [x=itr, y=t2, col sep=comma, color=green] {data/new-toy/nonunif-reset-slow-WeightedLossB600.csv};
     \addlegendentry{Nonuniform Mixture \\ ~[Ablation]}
    \addplot+[nonparam]         table [x=itr, y=t2, col sep=comma, color=brown] {data/new-toy/nonp-WeightedLossB-Adj600.csv};
    \addlegendentry{\Gls{np} \\ Mixture [Ours]}

    \addplot+[switch, highlight=0:4000]     coordinates {(4000, \plotbymin) (4000, \plotbymax)};
    \addplot+[switch, highlight=7000:9500] coordinates {(7000, \plotbymin) (7000, \plotbymax)};

    \nextgroupplot[
        ylabel={Task 3 \\ (Sawtooth)},
        xlabel={Meta-training iteration},
        ymin=\plotcymin, ymax=\plotcymax,
        xtick={0,1000,2000,3000,4000,5000,6000,7000,8000,9000,10000},
        xticklabels={$0$,$1000$,$2000$,$3000$,$4000$,$5000$,$6000$,$7000$,$8000$,$9000$,$10000$},
        legend pos=outer north east,
        ytick={30,40,50},
    ]

    \addplot+[unif-noreset]       table [x=itr, y=t3, col sep=comma, color=blue] {data/new-toy/unif-reset-slow-AvgLossB600.csv};
    \addplot+[nonunif-noreset]    table [x=itr, y=t3, col sep=comma, color=green] {data/new-toy/nonunif-reset-slow-WeightedLossB600.csv};
    \addplot+[nonparam]         table [x=itr, y=t3, col sep=comma, color=brown] {data/new-toy/nonp-WeightedLossB-Adj600.csv};

    \addplot+[switch, highlight=0:7000] coordinates {(4000, \plotcymin) (4000, \plotcymax)};
    \addplot+[switch]                   coordinates {(7000, \plotcymin) (7000, \plotcymax)};

  \end{groupplot}

  \node[label={[rotate=90]\textbf{\normalsize{Meta-test loss}}}] (title) at ($(group c1r2.center)+(-17.5em,0)$) {};

\end{tikzpicture}%

%% file: figures/psSW-weight.tex
\begin{tikzpicture}[baseline]

\begin{groupplot}[
    continuallearning,
    enlarge y limits,
    group style={group size=1 by 3, vertical sep=.5em},
    xmin=-1, xmax=9500,
    ymin=0.0, ymax=1.0,
    ytick={0.5,1.0},
    restrict y to domain*=0:80,  
    xticklabels={,,},
    height=0.95*\plotheight,
]

    \nextgroupplot[
      ylabel={Task 1 \\ (Polynomial)},
      ylabel style={align=center},
        xlabel={\hspace{20pt} \scriptsize{Phase 1 (Poly)} \hspace{58pt} \scriptsize{Phase 2 (Sinusoid)} \hspace{30pt} \scriptsize{Phase 3 (Sawtooth)}},
      xlabel shift={-52pt},
    ]

    \addplot+[cluster-1] table [x=itr, y=t1c1, col sep=comma] {data/new-toy/nonp-WeightsB-Og600.csv};
    \addplot+[cluster-2] table [x=itr, y=t1c2, col sep=comma] {data/new-toy/nonp-WeightsB-Og600.csv};
    \addplot+[cluster-3] table [x=itr, y=t1c3, col sep=comma] {data/new-toy/nonp-WeightsB-Og600.csv};


    \addplot+[switch, highlight=4000:9500] coordinates {(4000, 0) (4000, 1)};
    \addplot+[switch]                       coordinates {(7000, 0) (7000, 1)};

    \nextgroupplot[
      ylabel={Task 2  \\ (Sinusoid)},
      ylabel style={align=center},
      legend pos=outer north east,
    ]

    \addplot+[cluster-1] table [x=itr, y=t2c1, col sep=comma] {data/new-toy/nonp-WeightsB-Og600.csv};
    \addlegendentry{Cluster 1}
    \addplot+[cluster-2] table [x=itr, y=t2c2, col sep=comma] {data/new-toy/nonp-WeightsB-Og600.csv};
    \addlegendentry{Cluster 2}
    \addplot+[cluster-3] table [x=itr, y=t2c3, col sep=comma] {data/new-toy/nonp-WeightsB-Og600.csv};
    \addlegendentry{Cluster 3}


    \addplot+[switch, highlight=0:4000]     coordinates {(4000, 0) (4000, 1)};
    \addplot+[switch, highlight=7000:9500] coordinates {(7000, 0) (7000, 1)};

    \nextgroupplot[
      ylabel={Task 3  \\  (Sawtooth)},
      ylabel style={align=center},
      xlabel={Meta-training iteration number},
      xtick={0,1000,2000,3000,4000,5000,6000,7000,8000,9000,10000},
      xticklabels={$0$,$1000$,$2000$,$3000$,$4000$,$5000$,$6000$,$7000$,$8000$,$9000$,$10000$},
    ]

    \addplot+[cluster-1] table [x=itr, y=t3c1, col sep=comma] {data/new-toy/nonp-WeightsB-Og600.csv};
    \addplot+[cluster-2] table [x=itr, y=t3c2, col sep=comma] {data/new-toy/nonp-WeightsB-Og600.csv};
    \addplot+[cluster-3] table [x=itr, y=t3c3, col sep=comma] {data/new-toy/nonp-WeightsB-Og600.csv};


    \addplot+[switch, highlight=0:7000] coordinates {(4000, 0) (4000, 1)};
    \addplot+[switch]                   coordinates {(7000, 0) (7000, 1)};

  \end{groupplot}

  \node[label={[rotate=90]\textbf{\normalsize{Meta-test responsibility}}}] (title) at ($(group c1r2.center)+(-17.5em,0)$) {};

\end{tikzpicture}%

%% file: sections/07_literature_review.tex
\section{Related Work}
\label{sec:lit-rev}

\vspace{-4pt}

\paragraph{Meta-learning.}
In this work, we show how changes to the hierarchical Bayesian model assumed in meta-learning~\cite[][Fig.~1(a)]{grant2018recasting} can be realized as changes to a meta-learning algorithm. 
In contrast, follow-up approaches to improving the performance of meta-learning algorithms~\citep[\eg][]{lee2018gradient,finn2018probabilistic,gordon2019meta}
do not change the underlying probabilistic model; what differs is the inference procedure to infer values of the global (shared across tasks) and local (task-specific) parameters; for example, \cite{gordon2019meta} consider feedforward conditioning while \cite{finn2018probabilistic} employ variational inference.
Due to consolidation into one set of global parameters shared uniformly across tasks, none of these methods inherently accommodate heterogeneity or non-stationarity.

\paragraph{Continual learning.}
Techniques developed  to address the catastrophic forgetting problem in continual learning, such as \glsfirst{ewc}~\citep{kirkpatrick2017overcoming}, \glsfirst{si}~\citep{zenke2017continual}, \glsfirst{vcl}~\citep{nguyen2017variational}, and online Laplace approximation~\citep{ritter2018online} require access to an explicit delineation between tasks that acts as a catalyst to grow model size, which we refer to as \gls{taskaw}.
In contrast, our \gls{np} algorithm tackles the \gls{taskag} setting in which the meta-learner recognizes a latent shift in the task distribution and adapts accordingly.

%% file: figures/classification.tex
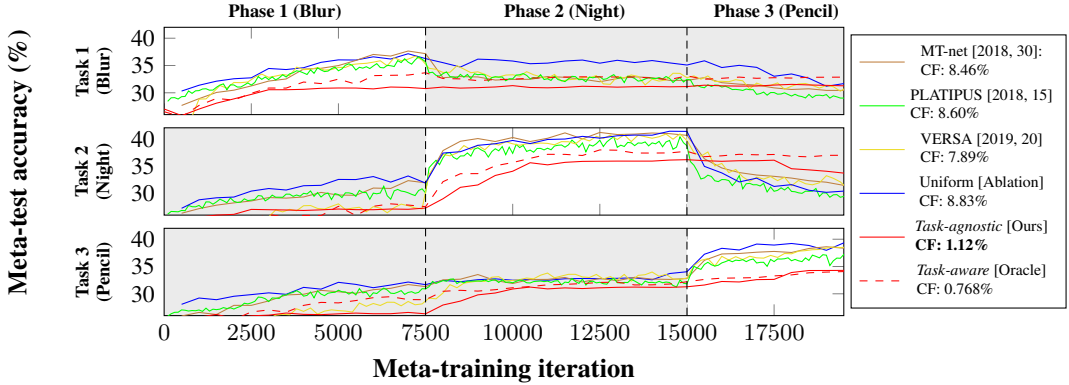
\begin{figure*}[t]
  \centering
  \input{figures/class-acc}
  \caption{
    Results on the evolving dataset of filtered \gls{mini} few-shot classification tasks (higher is better).
    Each panel (row) presents, for a specific task type (filter),
    the average meta-test set accuracy over cumulative number of few-shot episodes.
    We additionally report the degree of loss in
    backward transfer (catastrophic forgetting, \textbf{CF}) in the legend.
    This is calculated for each method as the average drop in accuracy on the first two tasks at the end of training
    (lower is better).
  }
  \label{fig:classification-accuracy}
\end{figure*}
\vspace{-16pt}
\begin{figure*}[t]
    \centering
    \input{figures/class-weight}
    \caption{
        Each panel (row) presents task-specific per-cluster meta-test responsibilities $\gamma^{(\ell)}$ 
        over time. A higher responsibility entails a higher degree of specialization of 
        a particular cluster (color) to a particular task (row).
    }
    \label{fig:bnp-weight}
\vspace{-15pt}
\end{figure*}

%% file: figures/class-acc.tex
\begin{tikzpicture}[baseline]


\newcommand{\plotymax}{.42}
\newcommand{\plotymin}{.26}

\begin{groupplot}[
      continuallearning,
      group style={group size=1 by 3, vertical sep=.5em},
      xmin=-1, xmax=19500,
      xtick={0,2500,5000,7500,10000,12500,15000,17500},
      ytick={.3,.35,.4},
      yticklabels={30,35,40},
      xticklabels={,,},
      ymin=\plotymin, ymax=\plotymax,
      restrict y to domain*=0:1,  
    ]

    \nextgroupplot[
        ylabel={Task 1 \\ (Blur)},
        xlabel={\hspace{20pt} \scriptsize{Phase 1 (Blur)} \hspace{58pt} \scriptsize{Phase 2 (Night)} \hspace{30pt} \scriptsize{Phase 3 (Pencil)}},
        xlabel shift={-52pt},
    ]


    \addplot+[unif-noreset] table [x=itr, y=t1, col sep=comma] {data/bnp-unif.csv};
    \addplot+[mt-net] table [x=itr, y=t1, col sep=comma] {data/bnp-mt.csv};
    \addplot+[versa] table [x=itr, y=t1, col sep=comma] {data/bnp-versa.csv};
    \addplot+[platipus] table [x=itr, y=t1, col sep=comma] {data/bnp-plat.csv};
    \addplot+[nonparam] table [x=itr, y=t1, col sep=comma] {data/aws_nbp/bnp_tagnostic.csv};
    \addplot+[nonparam, dashed] table [x=itr, y=t1, col sep=comma] {data/aws_nbp/bnp_taware.csv};

    \addplot+[switch, highlight=7500:19500]  coordinates {(7500, \plotymin) (7500, \plotymax)};
    \addplot+[switch]                        coordinates {(15000, \plotymin) (15000, \plotymax)};

    \nextgroupplot[
        ylabel={Task 2 \\ (Night)},
        legend style={align=left,at={(1.01,0.5)},anchor=west},
    ]
     \addplot+[mt-net] table [x=itr, y=t2, col sep=comma] {data/bnp-mt.csv};
     \addlegendentry{\acrshort{mt-net} \citep[2018,][]{lee2018gradient}: \\ {CF: 8.46\%} }

     \addplot+[platipus] table [x=itr, y=t2, col sep=comma] {data/bnp-plat.csv};
     \addlegendentry{\acrshort{platipus} \citep[2018,][]{finn2018probabilistic} \\ {CF: 8.60\%}}

     \addplot+[versa] table [x=itr, y=t2, col sep=comma] {data/bnp-versa.csv};
     \addlegendentry{\acrshort{versa} \citep[2019,][]{gordon2019meta} \\ {CF: 7.89\%}}

     \addplot+[unif-noreset] table [x=itr, y=t2, col sep=comma] {data/bnp-unif.csv};
     \addlegendentry{Uniform~[Ablation] \\ {CF: 8.83\%}}

     \addplot+[nonparam] table [x=itr, y=t2, col sep=comma] {data/aws_nbp/bnp_tagnostic.csv};
     \addlegendentry{\Gls{taskag}~[Ours] \\ \textbf{CF: 1.12\%}}

     \addplot+[nonparam, dashed] table [x=itr, y=t2, col sep=comma] {data/aws_nbp/bnp_taware.csv};
     \addlegendentry{\Gls{taskaw}~[Oracle] \\ {CF: 0.768\%}}

    \addplot+[switch, highlight=0:7500]     coordinates {(7500, \plotymin) (7500, \plotymax)};
    \addplot+[switch, highlight=15000:19500] coordinates {(15000, \plotymin) (15000, \plotymax)};

    \nextgroupplot[
        ylabel={Task 3 \\ (Pencil)},
        xlabel={Meta-training iteration},
        xtick={0,2500,5000,7500,10000,12500,15000,17500},
        xticklabels={$0$,$2500$,$5000$,$7500$,$10000$,$12500$,$15000$,$17500$},
    ]

    \addplot+[unif-noreset] table [x=itr, y=t3, col sep=comma]
    {data/bnp-unif.csv};

     \addplot+[mt-net] table [x=itr, y=t3, col sep=comma]
    {data/bnp-mt.csv};

    \addplot+[versa] table [x=itr, y=t3, col sep=comma]
    {data/bnp-versa.csv};

    \addplot+[platipus] table [x=itr, y=t3, col sep=comma]
    {data/bnp-plat.csv};


     \addplot+[nonparam] table [x=itr, y=t3, col sep=comma] {data/aws_nbp/bnp_tagnostic.csv};

    \addplot+[nonparam, dashed] table [x=itr, y=t3, col sep=comma] {data/aws_nbp/bnp_taware.csv};

    \addplot+[switch, highlight=0:15000] coordinates {(7500, \plotymin) (7500, \plotymax)};
    \addplot+[switch]                   coordinates {(15000, \plotymin) (15000, \plotymax)};

  \end{groupplot}

  \node[label={[rotate=90]\textbf{\normalsize{Meta-test accuracy (\%)}}}] (title) at ($(group c1r2.center)+(-17.5em,0)$) {};

\end{tikzpicture}%

%% file: figures/class-weight.tex
\begin{tikzpicture}[baseline]

\begin{groupplot}[
    continuallearning,
    enlarge y limits,
    group style={group size=1 by 3, vertical sep=.5em},
    xmin=-1, xmax=19500,
    ymin=0.0, ymax=1.0,
    ytick={0.5,1.0},
    restrict y to domain*=0:1,  
    xticklabels={,,},
    height=0.95*\plotheight,
]

    \nextgroupplot[
      ylabel={Task 1 \\ (Blur)},
      ylabel style={align=center},
      xlabel={\hspace{20pt} \scriptsize{Phase 1 (Blur)} \hspace{58pt} \scriptsize{Phase 2 (Night)} \hspace{30pt} \scriptsize{Phase 3 (Pencil)}},
      xlabel shift={-52pt},
    ]

    \addplot+[cluster-1] table [x=itr, y=t1c1, col sep=comma] {data/aws_nbp/bnp_weights.csv};
    \addplot+[cluster-2] table [x=itr, y=t1c2, col sep=comma] {data/aws_nbp/bnp_weights.csv};
    \addplot+[cluster-3] table [x=itr, y=t1c3, col sep=comma] {data/aws_nbp/bnp_weights.csv};


    \addplot+[dashed, switch, highlight=7500:19500]  coordinates {(7500, 0) (7500, 1)};
    \addplot+[dashed, switch]                        coordinates {(15000, 0) (15000, 1)};

    \nextgroupplot[
      ylabel={Task 2  \\ (Night)},
      ylabel style={align=center},
      legend style={align=center,at={(1.01,0.5)},anchor=west},
    ]

    \addplot+[cluster-1] table [x=itr, y=t2c1, col sep=comma] {data/aws_nbp/bnp_weights.csv}; \addlegendentry{Cluster 1}
    \addplot+[cluster-2] table [x=itr, y=t2c2, col sep=comma] {data/aws_nbp/bnp_weights.csv}; \addlegendentry{Cluster 2}
    \addplot+[cluster-3] table [x=itr, y=t2c3, col sep=comma] {data/aws_nbp/bnp_weights.csv}; \addlegendentry{Cluster 3}


    \addplot+[dashed, switch, highlight=0:7500]     coordinates {(7500, 0) (7500, 1)};
    \addplot+[dashed, switch, highlight=15000:19500] coordinates {(15000, 0) (15000, 1)};

    \nextgroupplot[
      ylabel={Task 3  \\ (Pencil)},
      ylabel style={align=center},
      xlabel={Meta-training iteration},
      xtick={0,2500,5000,7500,10000,12500,15000,17500},
      xticklabels={$0$,$2500$,$5000$,$7500$,$10000$,$12500$,$15000$,$17500$},
    ]

    \addplot+[cluster-1] table [x=itr, y=t3c1, col sep=comma] {data/aws_nbp/bnp_weights.csv};
    \addplot+[cluster-2] table [x=itr, y=t3c2, col sep=comma] {data/aws_nbp/bnp_weights.csv};
    \addplot+[cluster-3] table [x=itr, y=t3c3, col sep=comma] {data/aws_nbp/bnp_weights.csv};


    \addplot+[dashed, switch, highlight=0:15000] coordinates {(7500, -1) (7500, 2)};
    \addplot+[dashed, switch]                   coordinates {(15000, -1) (15000, 2)};

  \end{groupplot}

  \node[label={[rotate=90]\textbf{\normalsize{Meta-test responsibility}}}] (title) at ($(group c1r2.center)+(-17.5em,0)$) {};

\end{tikzpicture}%

%% file: sections/08_conclusion.tex
\section{Conclusion}
\label{sec:concl}

Meta-learning is a source of learned inductive bias.
Occasionally, this inductive bias is harmful because the experience
gained from solving a task does not transfer.
Here, we present an approach that allows a probabilistic meta-learner to explicitly 
modulate the amount of transfer between tasks, as well as to adapt its parameter 
dimensionality when the underlying task distribution evolves.
We formulate this as probabilistic inference in a mixture model that defines a
clustering of task-specific parameters.
To ensure scalability, we make use of the recent connection
between gradient-based meta-learning and hierarchical Bayes~\citep{grant2018recasting}
to perform approximate \glsfirst{map} inference 
in both a finite and an infinite mixture model.
Our work is a first step towards more realistic settings of
diverse task distributions, and crucially, \gls{taskag} continual meta-learning.
The approach stands to benefit from orthogonal improvements in posterior inference
beyond \gls{map} estimation (\eg
\acrlong{vi}~\citep{jordan1999introduction},
Laplace approximation~\citep{mackay1992practical},
or stochastic gradient \acrlong{mcmc}~\citep{metropolis1949monte}).
as well as scaling up the
base model (\eg trading the four-layer convolutional network for a more complex
architecture).

%% file: sections/09_supp.tex
\section{Extended related work}
\label{sec:supp}

\paragraph{Multi-task learning.} 
\cite{rosenstein2005transfer} demonstrated that negative transfer can worsen generalization performance, and avoidance of negative transfer has motivated much work on hierarchical Bayes in \gls{tl} and \gls{da}~\citep[\eg][]{lawrence2004learning,
yu2005learning,gao2008knowledge,daume2009bayesian,wan2012sparse}. 
Closest to our proposed approach is early work on hierarchical Bayesian multi-task learning with neural networks that places a prior only on the output layer~\citep{heskes1998solving, bakker2003task, salakhutdinov2013learning, srivastava2013discriminative}.
In contrast, we place a \gls{np} prior on the full set of neural network weights.
Furthermore, none of these approaches were applied to the episodic training setting of meta-learning. 
Similar to our point estimation procedure, \cite{heskes1998solving} and \cite{srivastava2013discriminative} propose training a mixture model over the output layer weights of a neural network using MAP inference. 
However, these approaches do not scale well to all the layers in a network as performing full passes on the dataset for inference of the full set of weights is computationally intractable in general.

\paragraph{Clustering.} 
Incremental or stochastic clustering was considered in the \gls{em} setting in \cite{neal1998view}.
and in the $K$-means setting in \cite{sculley2010web}.
\cite{lin2013online} conducted online learning of a \gls{np} mixture model using sequential variational inference. 
A key distinction between our work and these approaches is that we leverage the 
connection between \gls{eb} in a hierarchical model and \gls{gb} \gls{ml}~\citep{grant2018recasting} to use a
\Gls{maml}-like~\citep{finn2017model} objective as a log posterior surrogate.
This allows our algorithm to make use of a scalable stochastic gradient descent optimizer instead of alternating a maximization step with an inference pass over the full dataset~\citep[\cf][]{srivastava2013discriminative, bauer2017discriminative}. 

Our approach is also distinct from recent work on gradient-based clustering~\citep{greff2017neural} since we employ the episodic batching of \cite{vinyals2016matching}.
This can be a challenging setting for a clustering algorithm, as the assignments need to be computed using, for example, $K = 1$ examples per class in the 1-shot setting.

\paragraph{Contrasting the batch and stochastic settings.}
In the stochastic setting, access to past data is unavailable, 
and so none of the standard algorithms and heuristics for inference in \gls{np} models are 
applicable~\cite[\eg][]{jain2004split,hughes2012effective}.
In particular, our proposed algorithm does not refine the cluster assignments of previously observed points by way of multiple
expensive passes over the whole dataset.

In contrast, we incrementally infer model parameters and add components during episodic
training based on noisy estimates of the gradients of the marginal log-likelihood.
Moreover, we avoid the need to preserve task assignments, which is potentially harmful 
due to stale parameter values, since the task assignments in our framework are 
meant to be easily reconstructed on-the-fly
using the \estep with updated parameters $\metaparams\clusteridx{0}, \dots, 
\metaparams\clusteridx{\clustercount}, G$. 

\paragraph{\Acrlong{map} estimation as \acrlong{icm}.}

Due to the high-dimensionality of the parameter set of neural networks,
we consider a mode estimation procedure based on
\glsfirst{icm}~\citep{besag1986statistical,zhang2001segmentation,welling2006bayesian,raykov2016k}
that can leverage gradient computation instead of the expensive process of Gibbs sampling.
\gls{icm} is a greedy strategy that iteratively maximizes the full conditional
distribution for each variable (\ie computes the \gls{map} estimate), instead of sampling from the conditional
as is done in Gibbs sampling~\citep{welling2006bayesian}.
This leads to a fast point-estimation of the \gls{dpmm} 
parameters in which we only need to track the means of the cluster priors.

\paragraph{Alternative inference procedures in probabilistic mixtures.}
A standard approach for estimation in latent variable models, such as 
probabilistic mixtures, is to represent the distribution using samples produced via some sampling algorithm.
The most widely used is the Gibbs sampler~\citep{neal2000markov,gershman2012tutorial},
which draws from the conditional distribution of each latent variable, given the 
others, until convergence to the posterior distribution over all the latents.
However, in the setting of latent variables defined over high-dimensional parameter spaces
such as those of neural network models, using
a sampling algorithm such as the Gibbs sampler is prohibitively 
expensive~\citep{neal2012bayesian,muller1998issues}.
Instead of sampling, one can fit factorized variational distributions to the exact distribution $p(\phi, z|x) \approx q(\phi)q(z)$ ~\citep{ghahramani2000variational, blei2006variational}.
It should be noted that we do not claim that our method 
of point estimation in the \gls{dpmm} is the most accurate method for posterior inference
but we leave improved approximate inference extensions
to future work. 

The main drawback of using point estimates for a \gls{np} mixture estimation is the inability
to leverage the diffusion of the global prior $\globalprior$ when computing the likelihood of a new cluster.
Highly concentrated parameter estimates for non-empty clusters should lead to low likelihoods for outlier tasks,
whereas the diffused global prior should be better at capturing a wider variety of tasks.
Nonetheless, point estimation is a necessary trade-off between computation and accuracy.
To allow for a more accurate estimate of the likelihood, we experimented with simulating a normal
centered at the global prior mean with a variance hyperparameter that can be annealed over
time to account for increased certainty about the prior choice. We can then compare the average
cluster responsibility to the threshold.
Another interesting extension we experimented with was to compute the gradient for each of the samples
and average over the number of samples as to approximate the expectation of the gradient under
the global prior. However, we found this to be less stable than simply comparing the cluster
responsibilities to the threshold.

\section{\Acrlong{map} estimation in the \acrlong{dpmm}}
\label{sec:map-dpmm}

From (\ref{eq:crp-prior}) and using a conditional mode estimate for task-specific parameters
$\taskparams\taskidx{\taskidxvbl}$,
\begin{align}
\label{eq:map-dpmm}
    \log \prob{
        \clusterid\taskidx{\taskidxvbl} = \clusteridxvbl
        \given 
        \datapoint\taskidx{j\datapointidx{1:\datapointcountval}},
        \clusterid\taskidx{1:\taskidxvbl-1},
        \metaparams\clusteridx{\clusteridxvbl}
    } 
    \approx
    \left\{
        \begin{array}{ll}
            \pbox{12em}{$\log \observationcount\clusteridx{\clusteridxvbl} + 
            \log p(
            \datapoint\taskidx{j\datapointidx{1:\datapointcountval}}
            | \hat{\taskparams}\taskidx{\taskidxvbl}\clusteridx{\clusteridxvbl}) 
            +\log p( \hat{\taskparams}\taskidx{\taskidxvbl}\clusteridx{\clusteridxvbl} | \metaparams\clusteridx{\clusteridxvbl})$}   &
            \quad \textrm{for} \ \clusteridxvbl \leq \clustercount \\[1.3em]
            \pbox{12em}{$\log \concentration + 
            \log p(
            \datapoint\taskidx{j\datapointidx{1:\datapointcountval}}
            | \hat{\taskparams}\taskidx{\taskidxvbl}\clusteridx{\clusteridxvbl}) +
            \log (\hat{\taskparams}\taskidx{\taskidxvbl}\clusteridx{\clusteridxvbl} | \metaparams\clusteridx{0})$}  &
            \quad \textrm{for} \ \clusteridxvbl = \clustercount + 1
            ~.
        \end{array}
    \right.
\end{align}

\section{Experimental setup}

\subsection{Dataset details}

\paragraph{Few-shot regression}
\begin{itemize}
    \item Polynomial wave (\figref{polynomial}): $$y = \sum_i a_i x^{p_i}$$ and $a \sim \mathcal{U}(-5.0, 5.0)$.
    \item Sinusoid wave (\figref{sinusoid}): $$y = a \sin(x - \phi)$$ where $\phi \sim \mathcal{U}(0, \pi)$ and $a \sim \mathcal{U}(0.1, 5.0)$.
    \item Sawtooth wave (\figref{sawtooth}): $$y = -\frac{2a}{\pi}\arctan(\cot(\frac{x \pi}{\phi}))$$ where $\phi \sim \mathcal{U}(0, \pi)$, $a \sim \mathcal{U}(0.1, 5.0)$.
\end{itemize} 

\subsection{Hyperparameter choices}

\subsubsection{\Gls{mini} few-shot classification.}
\label{sec:supp-classification}

We use the same data split, neural network architecture, and hyperparameter values as in~\cite{finn2017model}
for common components.
We use $\tau=1$ for the softmax temperature and the same initialization as~\cite{finn2017model} 
for the global prior $\globalprior$.
We determine an iteration number for early stopping using the validation set.

\subsubsection{Continual few-shot regression.}
\label{sec:supp-continual-regression}
Our architecture is a feedforward neural network with $2$ hidden layers with ReLU nonlinearities, each of size $40$.
We use a meta-batch size of $10$ tasks (both for the inner updates and the meta-gradient updates) for $5$-shot regression.
Our \gls{np} algorithm starts with a single cluster
($\clustercount_0 = 1$ in \algref{em-dpmm}).
In these experiments, we set the spawning threshold $\spawningthreshold = 0.95 T/(\clustercount+1)$,
with $\clustercount$ the number of
non-empty clusters and $T$ the size of the meta-batch.
We use the mean-squared error for each task as the inner loop and
meta-level objectives.

\subsubsection{Continual few-shot \gls{mini} classification.}
\label{sec:supp-continual-classification}
We use the same data split, neural network architecture, and hyperparameter values as in~\cite{finn2017model}
for common components.
We use a meta-batch size of $4$ tasks, start with a single cluster, and set the spawning threshold to the same formula as in \secref{supp-continual-regression}.
We use the multi-class cross entropy error for each task as the inner loop and
meta-level objectives.
More details on the the practical implementation for image datasets of the \gls{np} algorithm can found in \secref{algorithm}.

\section{Practical and implementational details}
\label{sec:algorithm}

\subsection{\Gls{taskaw} \vs \gls{taskag}}
\label{sec:practical}

Since a cluster is not well-tuned immediately after its creation, we consider a 
cool-down period after the spawning of each new cluster where we do not consider the creation
of new clusters for a fixed number of iterations, and we freeze
the updating of existing clusters for a same number of iterations. 
This allows the newly-created 
cluster to take enough gradient updates in order to move from its global prior initialization, allowing
it to sufficiently differentiate from the global prior.

This experimental paradigm also allows us to approximate the \gls{taskaw} algorithms of prior work~\cite[\eg][]{kirkpatrick2017overcoming,zenke2017continual,nguyen2017variational,ritter2018online} which require access to an explicit delineation between tasks that acts as a catalyst to grow model size.
For the \gls{taskaw} \gls{np} mixture results reported in the experiments, we set this cool-down period to be exactly the length of the training phase for the appropriate dataset; therefore, clusters which are not meant to be specialized for the active dataset are not updated. In contrast, the \gls{taskaw} results consider a cool-down period of 1$k$ iterations, which is less than 15\% of the active period for each dataset. Extensions to this fixed cool-down period could consider the rate of learning in the active cluster in order to detect when the new component has been sufficiently fit to the new task.

\subsection{Practical extensions to the \gls{np} algorithm}
The penalty term of $\log \observationcount\clusteridx{\clusteridxvbl}$ or $\log \concentration$ is 
necessary to regularize the likelihood of a potential new cluster in order to limit overspawning.
However, in the setting where the likelihood is approximated by the loss function of a complex neural network,
as in the case for most meta-learning applications,
there is a large difference in orders of magnitude between the loss value 
(especially for the cross-entropy function) and the penalty term,
even after a single batch of assignments.
Furthermore, the classical log observation count $\log{n}$ term is misaligned with our stochastic 
setting for two reasons.
First, since we do not re-evaluate over the whole dataset for every meta-learning episode, we are thus more
concerned with the relative number of task assignments over recent iterations than the total number of
assignments over the duration of training.
Second, the number of tasks to be assigned can grow too large in the stochastic
setting (e.g. $60k$ for \gls{mini}) which exacerbates the already large difference in orders of magnitudes
between the loss function and the penalty term.

Accordingly, we propose two changes;
First, we compute the observation based on a moving window of fixed size ($5$ in the experiments).
Second, we apply a coefficient, which can be tuned, to the log observation count in (\ref{eq:crp-prior}).
This provides more flexibility to our meta-learner as it allows it to apply to any black-box function
approximator which might exhibit losses of orders of magnitudes smaller than those expected of classical
probabilistic models.
%
%
While the moving window size and CRP penalty coefficient terms are somewhat interdependent,
we propose them as a simple starting point to tune this \gls{np} meta-learner beyond what is empirically explored in this paper.

Note that without such changes in the stochastic setting of meta-learning, a nonparametric algorithm
would be unable to spawn a new cluster after the first handful of iterations.
Even if we were to lower the threshold $\epsilon$, multiple almost identical clusters would be spawned in
the first few iterations before it would be impossible to spawn anymore.
Furthermore, the clusters would be nearly identical given the small step size of a gradient
update for each meta-learning episode.
Finally, this would be computationally intensive since unlike the typical applications of
\gls{np} mixture learning where one can afford to spawn hundreds of components then
prune them over the training procedure.

\subsection{Thresholding}

A marked difference that is not immediate from the Gibbs conditionals
is the use of a threshold on the cluster responsibilities,
detailed in the \estep in Subroutine 4,
to account for noise from stochastic optimization when spawning a cluster on the basis of a single batch.
This threshold is necessary for the stochastic mode estimation procedure of \algref{em-dpmm}, as it ensures that a 
new cluster's responsibility needs to exceed a certain value before being permanently added to the set of 
components.

If a cluster has close to an equal share of responsibilities as compared to existing clusters after
accounting for the \gls{crp} penalty $\log n \clusteridx{\clusteridxvbl}$ or $\log \concentration$,
it is spawned.
Accordingly, this approximate inference routine 
still preserves the preferential attachment (``rich-get-richer'') dynamics of Bayesian nonparametrics~\citep{raykov2016k}. 
A sequential approximation for \gls{np} mixtures with a similar threshold
was proposed in \cite{lin2013online} and \cite{tank2015streaming}, in which variational Bayes was used 
instead of point estimation in a \gls{dpmm}.

\subsection{Pruning heuristics}
None of the results reported in our experiments used a pruning heuristic as we used a
rather conservative hyper parameter setting that deters overspawning. 
We did however explore different heuristics which could work in more general settings,
especially in the presence of many more latent clusters of tasks than considered in the experimental settings in this work.
One such heuristic is to prune small clusters that have received disproportionately few assignments 
over a certain number of past iterations.
Another is to evaluate the functional similarity of two clusters by computing an odds-ratio
statistic for the assignment probabilities to each cluster over a set of validation tasks.
If the odds-ratio statistic is below a certain threshold, the smaller cluster can be pruned.


\subsection{Estimating the CRP hyperparameters}
We fixed $\alpha$ at the size of the meta-batch. An alternative is to place a $\Gamma(1,1)$
on the concentration parameter. 
Based on the likelihood~\todocite{reference needed}, the posterior is then proportional to 
$p(\alpha | N, K) \propto \frac{\Gamma(\alpha)}{\Gamma(\alpha + N)} \alpha^{K} e^{-\alpha}$
This is not a standard distribution but \cite{rasmussen2000infinite} have shown that
$\log{p(\alpha | N, K)}$ is log-concave and methods such as L-BFGS have been used
successfully in prior works.
Alternatively, if we have some prior knowledge about the expected number of clusters,
we can compute $\alpha$ based on $E[K] = \alpha \log{N}$.
For the window-size, we considered an initial size of 20 iterations that can grow as more cluster
are considered.

%
\subsection{Implementation details}
We implemented both of our parametric and \gls{np} meta-learners in \gls{tf}~\cite{abadi2016tensorflow}.
We considered 2 different settings for the \mstep optimization: 
\begin{itemize}
	\item Train each cluster’s parameters separately based on its corresponding
	loss function in an alternating manner closest to the classic EM algorithm.
	\item Train all cluster weights simultaneously using a surrogate loss over
	all validation batches. 
\end{itemize}
Since the latter better leverages the differentiability of softmax-clustering and performed better empirically, 
we used it to report all experimental results.

\subsubsection{Nonparametric Implementation}
For the nonparametric algorithm, we chose the first approach to the \mstep by
constructing separate optimizers for each cluster's parameters.
We pre-allocate a set of weights and use a mask during training to discard
the parameters of empty clusters due to the static nature of \gls{tf} graphs.
When the algorithm exhausts the set of pre-allocated
weights, we simply construct more network weight and reinitialize our optimizers.

\subsubsection{\Gls{crp} global prior}

The likelihood of a new cluster is sensitive to the choice of a base measure or
prior prior, $\globalprior$ on the cluster hyperparameters.
Our gradient-based point estimation does not make any modeling assumption
on the distribution of the weights, rendering the problem of principally updating the
base measure, after or during training, non-trivial.
We chose to initialize all weights with zero-mean normals in the fully-connected
layers. For the convolutional layers, we leveraged Xavier initialization~\cite{glorot2010understanding}
similarly to prior work~\cite{finn2017model} in \gls{ml}.

However, such initialization is poor in the \gls{np} for most non-trivial regression or classification tasks.
Therefore, in the nonparametric setting, we start with a single cluster for a fixed number of
iterations. We then initialize all clusters with the weights of the first clusters.
This set of weights can be considered as the mean of the base measure or global prior in our setting.

We periodically update the global prior using a uniform average of the parameters of the existing clusters.
This can be done by simply averaging over the parameter of
the non-empty clusters as weighted by their sizes. Note that, 
we found that performing weighted KDE smoothing with a small bandwidth hyperparameter to
perform slightly better than the average which is to be expected for neural network parameters.
\ghassen{averaging is usually bad but not sure what to cite}
The number of iterations between updates of the global prior is a hyperparameter that we tune on the validation set.
It is also possible to continuously, but less frequently over time, update this global prior
 as more data is encountered. 